\documentclass[letterpaper,10pt,journal,twoside]{IEEEtran}
\usepackage{amsmath,amsfonts}
\usepackage{algorithmic}
\usepackage{array}
\usepackage{textcomp}
\usepackage{stfloats}
\usepackage{url}
\usepackage{verbatim}
\usepackage{graphicx}
\def\BibTeX{{\rm B\kern-.05em{\sc i\kern-.025em b}\kern-.08em
    T\kern-.1667em\lower.7ex\hbox{E}\kern-.125emX}}
\usepackage{balance}
\usepackage{graphics}
\usepackage{subfigure}
\usepackage{booktabs}
\usepackage{multirow}
\usepackage{cite}
\usepackage{tikz,xcolor,hyperref}
\usepackage{algorithm}
\usepackage{fontawesome}

\definecolor{lime}{HTML}{A6CE39}
\DeclareRobustCommand{\orcidicon}{%
	\begin{tikzpicture}
	\draw[lime, fill=lime] (0,0) 
	circle [radius=0.16] 
	node[white] {{\fontfamily{qag}\selectfont \tiny ID}};
	\draw[white, fill=white] (-0.0625,0.095) 
	circle [radius=0.007];
	\end{tikzpicture}
	\hspace{-2mm}
}

\foreach \x in {A, ..., Z}{%
	\expandafter\xdef\csname orcid\x\endcsname{\noexpand\href{https://orcid.org/\csname orcidauthor\x\endcsname}{\noexpand\orcidicon}}
}

\begin{document}
\title{EV-MGDispNet: Motion-Guided Event-Based Stereo Disparity Estimation Network with Left-Right Consistency}
\author{Junjie Jiang$^\dagger$\orcidA{}, Hao Zhuang$^\dagger$\orcidB{}, Xinjie Huang\orcidC{}, Delei Kong\orcidD{}, Zheng Fang\orcidF{}, \emph{Member, IEEE}
\thanks{
Manuscript received: August 1, 2024; Revised: August 1, 2024; Accepted: August 1, 2024. This work was supported in part by the National Natural Science Foundation of China under Grant 62073066 and Grant U20A20197, in part by the Fundamental Research Funds for the Central Universities under Grant N2226001, in part by the 111 Project under Grant B16009, and in part by the Intel Neuromorphic Research Community (INRC) under Grant RV2.137.Fang. \emph{(Corresponding author: Zheng Fang.)}

Junjie Jiang, Hao Zhuang, Xinjie Huang and Zheng Fang are with Faculty of Robot Science and Engineering, Northeastern University, Shenyang, China (e-mail: jianggalaxypursue@gmail.com, howechong9@gmail.com, gg.xinjie.huang@gmail.com, fangzheng@mail.neu.edu.cn), Delei Kong is with School of Robotics, Hunan University, Changsha, China (kong.delei.neu@gmail.com).

$\dagger$ These authors contributed equally.

Digital Object Identifier (DOI): see top of this page.
}}

\markboth{IEEE Transactions on Instrumentation and Measurement, Vol. ?, No. ?, July 2024}%
{Huang \MakeLowercase{\textit{et al.}}: EV-MGDispNet}

\maketitle

\begin{abstract}
Event cameras have the potential to revolutionize the field of robot vision, particularly in areas like stereo disparity estimation, owing to their high temporal resolution and high dynamic range.
Many studies use deep learning for event camera stereo disparity estimation.
However, these methods fail to fully exploit the temporal information in the event stream to acquire clear event representations. Additionally, there is room for further reduction in pixel shifts in the feature maps before constructing the cost volume.
In this paper, we propose EV-MGDispNet, a novel event-based stereo disparity estimation method.
Firstly, we propose an edge-aware aggregation (EAA) module, which fuses event frames and motion confidence maps to generate a novel clear event representation.
Then, we propose a motion-guided attention (MGA) module, where motion confidence maps utilize deformable transformer encoders to enhance the feature map with more accurate edges.
Finally, we also add a census left-right consistency loss function to enhance the left-right consistency of stereo event representation.
Through conducting experiments within challenging real-world driving scenarios, we validate that our method outperforms currently known state-of-the-art methods in terms of mean absolute error (MAE) and root mean square error (RMSE) metrics. 
\end{abstract}

\begin{IEEEkeywords}
Stereo disparity estimation, event camera, deep learning, left-right consistency, spatio-temporal feature enhancement.
\end{IEEEkeywords}

\section{Introduction}
\label{sec:introduction}

\begin{figure}[htbp]
\vspace{-0em}
\centering
\includegraphics[width=\columnwidth]{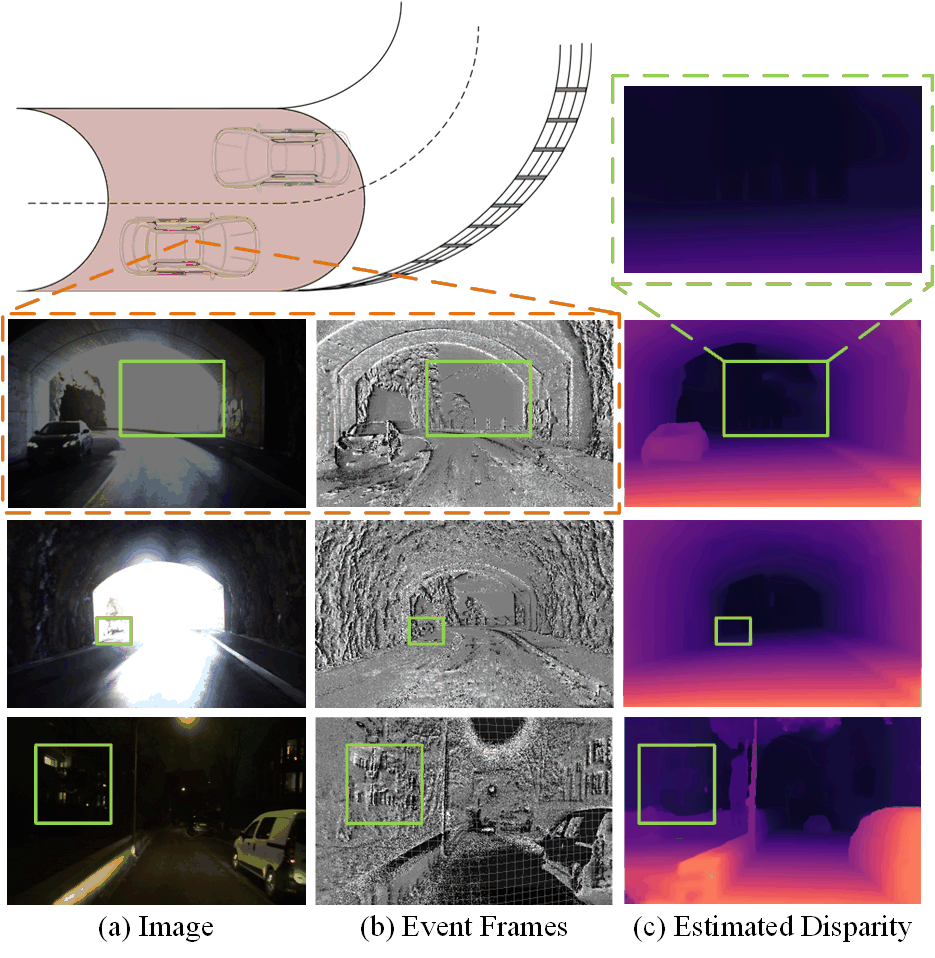}
\vspace{-2em}
\caption{The stereo disparity estimation results of event cameras at daytime tunnel exits and nighttime urban roads (extracted from the interlaken$\_$00$\_$b and zurich$\_$city$\_$12$\_$a test sequences from the DSEC dataset\cite{DSEC}) are evaluated. Our stereo disparity estimation method demonstrates good performance in these scenarios. (Event frames in (b) are aggregated edge-modulated event frames.)}
\label{fig:1}
\vspace{-1em}
\end{figure}

\IEEEPARstart{E}{vent} cameras \cite{gallego2020event,chen2020event,zheng2023deep,kong2022event,zhuang2023ev,wu2021novel}, inspired by the neural structure of the human eye in response to the changes of illumination, are a type of neuromorphic vision sensor\cite{dong2015review}. They possess the ability to asynchronously capture sub-microsecond brightness changes at each pixel, output pixel coordinates, timestamps and polarities of these brightness changes.
Event camera presents several advantages, including low latency, high temporal resolution, high dynamic range (HDR) and abundant motion information. These attributes make it exceptionally well-suited for visual tasks in scenes with high dynamic range and rapid motion. Considering the aforementioned advantages, it is highly worthwhile to investigate the application of event cameras in stereo depth estimation.
Stereo depth estimation\cite{yin2022ego, wei2022optimization} stands as a fundamental task in computer vision\cite{hong2021stereopifu} and robotics \cite{scharstein2002taxonomy}. 
It involves calculating the disparity of corresponding pixel points in a pair of images with a known pose relationship, which is also known as stereo disparity estimation. 
Subsequently, the depth is retrieved through a process known as triangulation. 
It possesses extensive applications in the tasks of robotics\cite{schmid2013stereo}, autonomous driving\cite{feng2022mafnet, sun2023framework} and virtual augmented reality 
 (VR/AR)\cite{valentin2018depth}, offering a broad range of possibilities.

Previous event-based stereo disparity estimation pipelines using deep learning \cite{tulyakov2019learning,mostafavi2021event,ahmed2021deep,nam2022stereo,zhang2022discrete,uddin2022unsupervised} make significant contributions to this emerging field, yielding outstanding outcomes.
Similar to frame-based stereo disparity learning\cite{tong2022adaptive,chong2023adaptive}, the methods proposed in the above works generally adhere to the following pipeline: First, stereo features are extracted through a feature extractor. Next, cost volumes are constructed and aggregated. Finally, the aggregated cost volumes are utilized for stereo disparity estimation.
However, unlike RGB frames, event data manifest as spatio-temporal point streams that are sparse in the pixel plane and temporally dense.
To be compatible with existing stereo disparity estimation frameworks, we need to transform event streams into image-like event representations.
A naive approach to represent event streams is to reconstruct the events into images\cite{ahmed2021deep,mostafavi2021event}, and yet the quality of reconstructed images significantly impacts the accuracy of stereo disparity estimation.
Thus, a more intuitionistic approach is to construct representations directly from event streams.
Hand-crafted event representations can preserve a greater amount of raw event data and exhibit more stable outcomes compared to reconstructed images\cite{tulyakov2019learning,ahmed2021deep}.
But, due to memory limitations, they often struggle to efficiently leverage event data over longer time windows, leading to insufficient accuracy in disparity estimation at the scene edges and object contours.
Learning-based event representation methods\cite{mostafavi2021event,nam2022stereo,zhang2022discrete} can overcome the aforementioned limitations, yielding more compact representations. However, existing approaches lack rich event timestamps associated with motion confidence, constraining the correctness of event representations in geometric structures and undermining the left-right consistency of the stereo camera.

Furthermore, event representation generally fails to effectively address the issue of pixel shifts in intermediate feature maps, i.e., the stacking of structural information of the same object at different time instances in the feature map.
In event-based stereo disparity estimation, pixel shifts in feature maps significantly impact the construction of cost volumes.
Previous works\cite{zhang2022discrete,ahmed2021deep} introduce conditional normalization to address this issue by adjusting the distribution of feature maps to better align with the edges of scenes.
However, conditional normalization relies on global adjustments to feature distribution, making it difficult to generate adjusted feature maps precisely align with real-world scenes. 
Therefore, these methods exhibit limited efficacy.

\begin{figure*}[htbp]
\vspace{-0.0em}
\centering
\includegraphics[width=\textwidth]{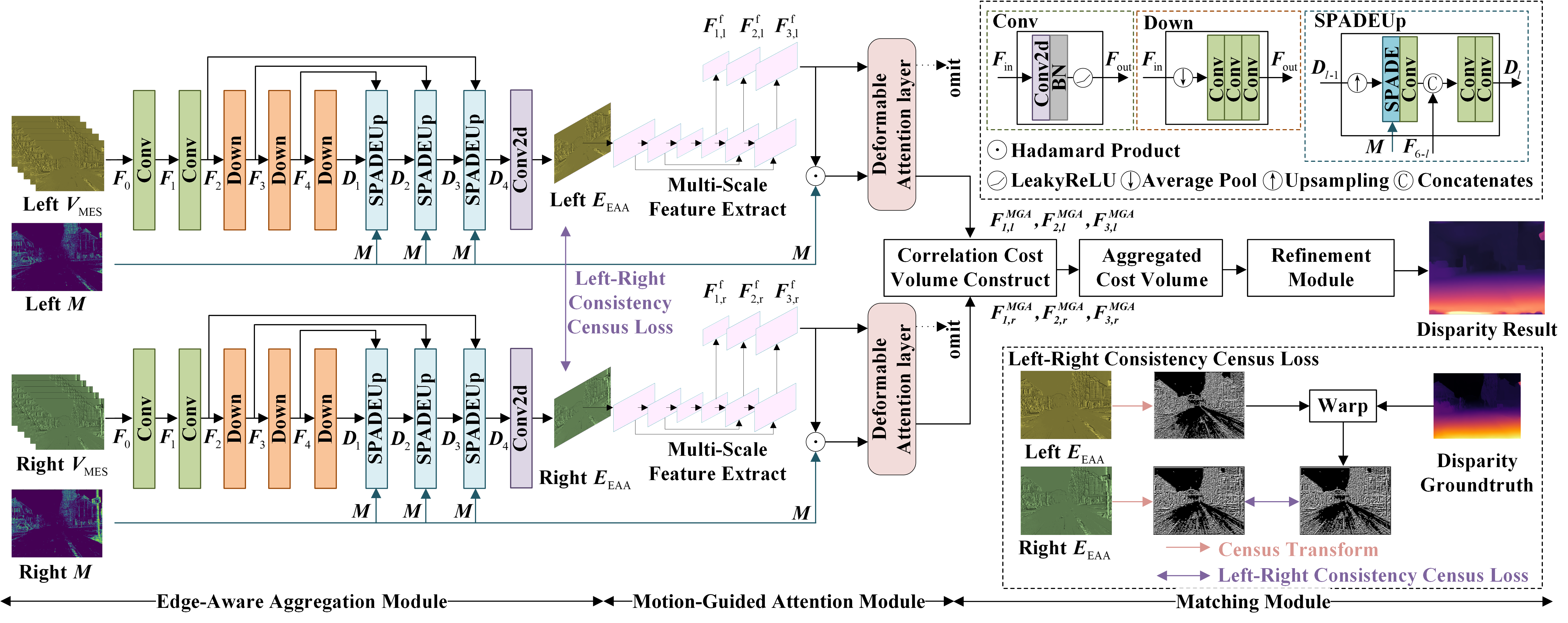}
\vspace{-2.0em}
\caption{
A detailed description of the proposed EV-MGDispNet pipeline is as follows. Firstly, the MES is used as input to the EAA module, which generates a clear aggregated edge-modulated event frame by fusing temporal information through motion confidence maps.
Afterwards, the MGA module is utilized to extract and integrate motion confidence maps with aggregated edge-modulated event frame features, yielding an edge-enhanced feature map.
Next, the cost volume is computed using the left-right edge-enhanced feature maps and aggregated using ISA and CSA modules. 
Finally, the refined module is used to obtain the stereo disparity estimation results. We use the smooth $L_{1}$ loss and left-right consistency census loss to train our network.}
\label{fig:2}
\vspace{-1.0em}
\end{figure*}

To sum up, previous works do not yet reach a consensus on addressing the challenges related to event representation and feature enhancement. To further resolve these issues, we propose EV-MGDispNet, a novel motion-guided event-based stereo disparity estimation network with left-right consistency, as shown in Fig. \ref{fig:2}.
It fuses temporal information through motion confidence maps to generate clear event representation and lower pixel shifts feature maps. 
We evaluate our method (Fig. \ref{fig:1}) on the DSEC dataset and experimental results show that our method achieves state-of-the-art (SOTA) performances in MAE and RMSE. 
In summary, the main contributions of this paper are as follows:

\begin{itemize}
\item We propose an edge-aware aggregation (EAA) module, composed of a convolutional encoder and a spatially adaptive denormalization (SPADE) upsampling module, which can fuse event frames information and motion confidence maps to generate new clear and more accurate event representation.
\item We propose a motion-guided attention (MGA) module, where motion confidence maps utilize deformable transformer encoders to guide the generation of aggregated edge-modulated event frame features, resulting in an enhanced feature map with more accurate edges, thereby reducing pixel offset in the feature maps, benefiting the construction of feature cost voxels and effectively improving the accuracy of stereo disparity estimation.
\item We propose a left-right consistency census loss to supervise the event representation output from the EAA module. This enhances the left-right consistency of the stereo event representation, making it more compliant with the stereo camera model.
\item We compare our proposed method with other SOTA methods on the DSEC dataset and demonstrate that our method has advanced performance.
\end{itemize}

The remainder of this article is structured as follows. Section \ref{sec:relatedwork} provides an overview of the related work concerning event-based stereo disparity estimation. Section \ref{sec:methodology} outlines the overall network pipeline of our EV-MGDispNet, offering detailed insights into event representation, edge-aware aggregation (EAA) module, motion-guided attention (MGA) module and left-right consistency census loss. Next, the experimental results of our EV-MGDispNet on the DSEC dataset are presented in Section \ref{sec:experiments}. Finally, Section \ref{sec:conclusions} concludes this article.

\section{Related Work}
\label{sec:relatedwork}
Nowadays, deep learning is widely applied in event-based stereo disparity estimation. Existing work can be succinctly described in two aspects: event representation and event-based stereo disparity estimation network architectures.

\textbf{Regarding event representation}. \cite{tulyakov2019learning,ahmed2021deep,uddin2022unsupervised} transforms events into an event queue, initially proposed by Tulyakov et al.\cite{tulyakov2019learning} , for stereo disparity estimation. The event queue is a 4D tensor that preserves both temporal and spatial information through a First-In-First-Out (FIFO), providing more adequate feature information for the left and right features after feature extraction. However, the temporal window size that this representation can preserve is dynamically constrained by the size of the FIFO. Unless the camera movement is exceedingly slow (in stationary scenes), it can only retain information for a relatively short period. This, in practice, diminishes the available event information that can be utilized.
Mostafavi et al.\cite{mostafavi2021event} proposes a method that combines event stream within a time window with images using image reconstruction to generate reconstructed images. Subsequently, these reconstructed left and right images are utilized for stereo disparity estimation. Although the reconstructed images contribute to enhancing the accuracy of disparity estimation, proficient reconstruction networks often come with higher computational demands.
Nam et al.\cite{nam2022stereo} proposes a learning-based approach to generate event representations by learning from multiple event frames within various time windows. This method of representation generating through learning reduces the computational demands on the event representation compared to image reconstruction methods. However, the way in which representations are generated does not incorporate actual temporal information, restricting the network's ability to understand the real edges and contours of objects in reality.
Zhang et al.\cite{zhang2022discrete} combine neuromorphic activation functions and convolution to fuse event frames from multiple fixed-time windows, generating event features. This approach enables the fusion of information from several event frames, somewhat retaining the temporal information within the event stream. However, the network still processes event frames from fixed-time windows as inputs, impacting the accuracy of edges and contours due to the fixed-time window size. Smaller time windows offer more accurate edges but contain less event information and require increased computational resources\cite{nam2022stereo}.

\textbf{Regarding network architectures}. Deep learning-based event stereo disparity estimation algorithms \cite{tulyakov2019learning,mostafavi2021event,ahmed2021deep,nam2022stereo,zhang2022discrete,uddin2022unsupervised} share the commonality of extracting event stream features through a network and expressing them as feature maps, and then using a frame-based stereo disparity estimation network to estimate disparity. \cite{tulyakov2019learning} proposes the earliest deep learning-based method for event stereo disparity estimation. They convert events into an event queue and extract features using a continuous fully-connected layer, and finally use the event features as input to a stereo disparity estimation network based on images to achieve dense disparity estimation. Although it contains temporal information, its ability to express local structure and edge accuracy is still insufficient. \cite{mostafavi2021event} proposes the earliest deep learning-based method that fuses event and frame image data for stereo disparity estimation. They use an iterative structure\cite{choi2020learning,mostafavi2021e2sri} to fuse information from multiple event frames and image frames, which reduces the efficiency of the network. \cite{ahmed2021deep} reconstructs event features as images and combines them with frame image features using the conditional normalization method. However, this brings potential risks. If the reconstruction quality is poor, it damages the network's performance and greatly increases the network's training and inference time. \cite{nam2022stereo} utilizes learning-based representation and employs knowledge transfer methods to learn a model that incorporates both past and future information. However, during the inference process, only past data is utilized to predict the future. They improve the disparity estimation performance using knowledge distillation\cite{deng2021learning,gao2020compact,ding2022kd,sun2022ess} by incorporating future event information to reinforce the disparity estimation results based on only past event information. \cite{zhang2022discrete} uses continuous-time convolution (CTC) and discrete-time convolution (DTC) to extract event features and uses the conditional normalization method to enhance the boundary effect of dense disparity. However, these two methods do not incorporate the timestamp information of events, and the network can only learn edge expression through the scene's shape, leading to incorrect geometric shapes in event features in many scenarios.

\section{Methodology}
\label{sec:methodology}
In this section, we present a detailed description of the methodology and design of our proposed EV-MGDispNet, as illustrated in Fig. \ref{fig:2}. We will describe each module component, main step and some considerations in the network training process in this part.

\subsection{The Overall Architecture}

Fig. \ref{fig:2} illustrates the overall architecture of our proposed EV-MGDispNet. Firstly, events are converted into mixed-density event stacking (MES)\cite{nam2022stereo}. Next, we use the EAA module  to fuse information from the motion confidence maps and MES, and transform MES into an aggregated edge-modulated event frame with richer details (Sec. \ref{spade-unet}). 
Afterwards, we utilize the motion-guided attention (MGA) module to extract and fuse motion confidence maps and aggregated edge-modulated event frame features to obtain the edge-enhanced feature map, it is then utilized to construct cost volume. 
Next, we construct a cost volume pyramid using the similarity (correlation) method and aggregate the cost volumes using the ISA and CSA\cite{xu2020aanet}.
Finally, we utilize the winner-take-all (WTA)\cite{scharstein2002taxonomy} and refinement module\cite{refinement} to estimate disparity.
In terms of the loss function, we utilize the smooth $L_{1}$ loss at the final layer of the network and the left-right consistency census loss after the EAA module.

\subsection{Event Camera and Event Representation}
\label{Event Representation}

The event camera simulates the functionality of the transient visual pathway of the biological retina, where each pixel independently and asynchronously responds to relative brightness changes with a time resolution at the microsecond-level\cite{lichtensteiner2008128x128,delbruck2014integration}. The triggering condition for each pixel is as follows:
\begin{equation}
\label{eq:1}
\left|\log{\boldsymbol{I}(\boldsymbol{x}_{i},t_{i})}-\log{\boldsymbol{I}(\boldsymbol{x}_{i},t_{i}-\Delta t_{i})}\right|\ge\mathrm{C},
\end{equation}
where $\boldsymbol{I}$ is the latent intensity on the pixel plane, $i$ is the index for the pixel at position $\boldsymbol{x}_{i}=(x_{i},y_{i})^\top$, $t_{i}$ is the time when pixel $\boldsymbol{x}_{i}$ is triggered, $t_{i}-\Delta t_{i}$ is the time when pixel $\boldsymbol{x}_{i}$ is last triggered, and $\mathrm{C}$ is the event triggering threshold. An event is triggered when the absolute value of the logarithmic intensity difference is greater than $\mathrm{C}$. The triggered pixel returns an event $\boldsymbol{e}_{i}=(x_{i},y_{i},t_{i},p_{i})^\top$, where $p_{i} \in \{-1,+1\}$ is the polarity of the event, which is $1$ when the logarithmic intensity $\log_{}{\boldsymbol{I}(\boldsymbol{x}_{i},t_{i})}$ at time $t_{i}$ is greater than the logarithmic intensity $\log{\boldsymbol{I}(\boldsymbol{x}_{i},t_{i}-\Delta t_{i})}$ at time $t_{i}-\Delta t_{i}$, and $-1$ otherwise.
Since a single event contains very little information and the event camera does not produce data when it is stationary, we use the MES \cite{nam2022stereo} as the input to our network. Specifically, it is expressed as follows:
\begin{equation}
\label{eq:2}
\boldsymbol{V}_\text{MES}(x,y,l) =\sum_{\boldsymbol{e}_{i} \in \boldsymbol{\varepsilon }_{\mathrm{N}_{e}/2^{l}}} p_{i} \boldsymbol{\delta}\left(x-x_{i}, y-y_{i}\right),
\end{equation}
where $\boldsymbol{V}_\text{MES}$ has a shape of $(\mathrm{H},\mathrm{W},\mathrm{L})$,  $l = \{0,1,2,3,...,\mathrm{L}-1\}$, $\boldsymbol{\varepsilon }_{\mathrm{N}_{e}/2^{l}}$ is the set of $\mathrm{N}_{e}/2^{l}$ event points accumulate from time $t_{i}$ backwards, and $\mathrm{N}_{e}$ is the total number of event points, $\delta(\cdot)$ is the dirac pulse.

\subsection{Edge-Aware Aggregation Module (EAA Module) }
\label{spade-unet}

\begin{figure}[htbp]
\vspace{-1.0em}
\centering
\includegraphics[width=0.7\columnwidth]{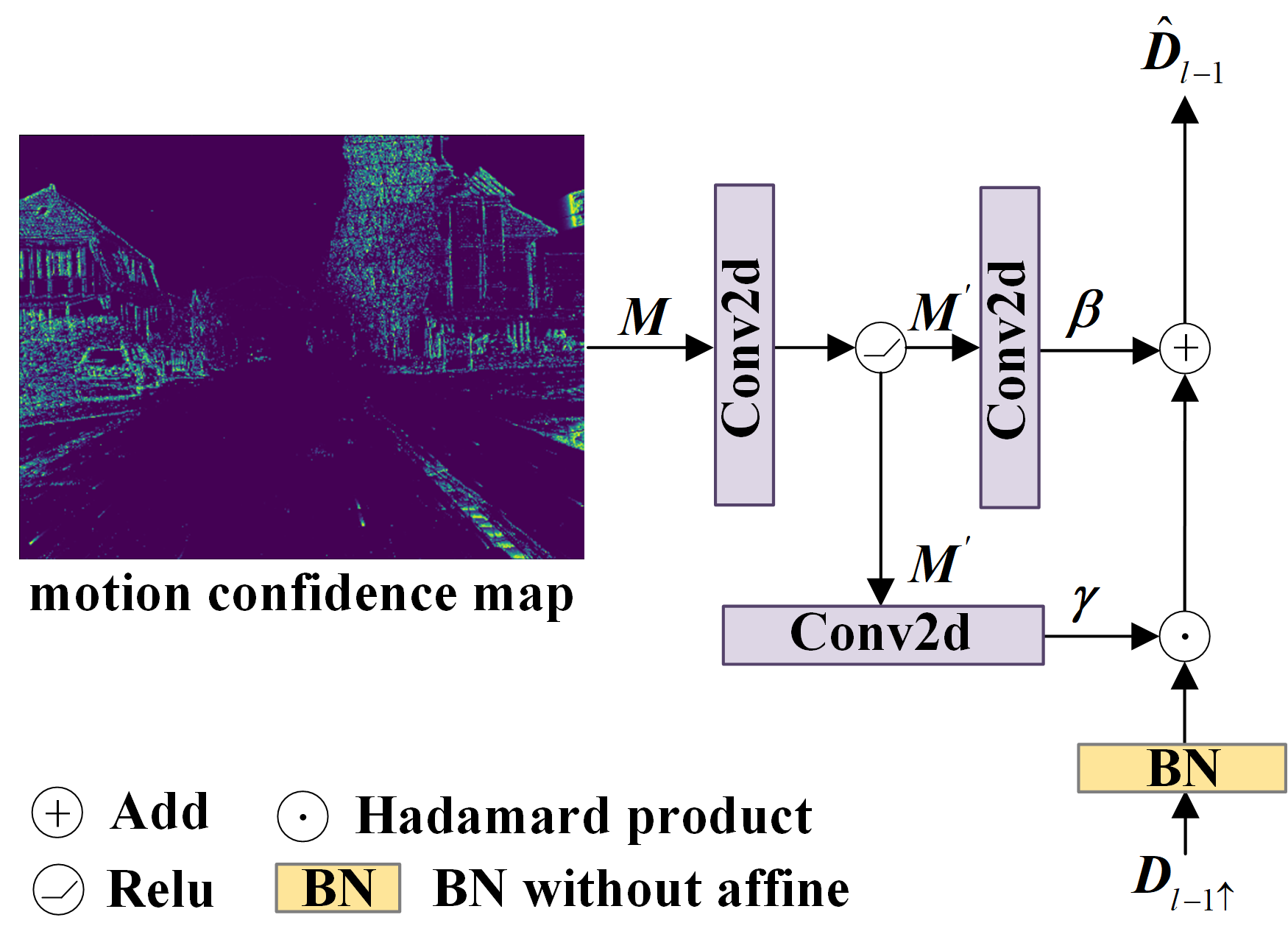}
\vspace{-1.0em}
\caption{Illustration of the SPADE layer structure is as follows. It utilizes motion confidence maps to obtain tensors $\boldsymbol{\gamma}$ and $\boldsymbol{\beta}$, which modulate the distribution of features $\boldsymbol{D}_{l-1\uparrow}$ from the decoder.}
\label{fig:3}
\vspace{-1.0em}
\end{figure}

In stereo disparity estimation, accurate disparity estimation relies on input images with clear object boundaries\cite{pan2021dual,kopf2021robust,gosta2010accomplishments}. 
Previous approaches attempt to generate clear event representations from event frames with different time windows\cite{nam2022stereo}.
Nonetheless, these edges are still inaccurate, because in the absence of temporal information, the network cannot know the true nature of object boundaries and texture within objects in reality. To address this issue, we propose an edge-aware aggregation (EAA) module, the implementation is detailed as follows.
Firstly, we use $\boldsymbol{V}_\text{MES}$ as the input of the EAA module and obtain the shallow feature $\boldsymbol{F}_1$, $\boldsymbol{F}_2$ through the convolution module $\boldsymbol{f}_\text{conv}(\cdot)$ to extract the shallow feature map without reducing its resolution, which can be expressed as follows:
\begin{equation}
\label{eq:3}
\begin{aligned}
\boldsymbol{F}_\text{0} &= \boldsymbol{V}_\text{MES}, \\
\boldsymbol{F}_\text{1} &= \boldsymbol{f}_\text{conv}(\boldsymbol{F}_\text{0}), \\
\boldsymbol{F}_\text{2} &= \boldsymbol{f}_\text{conv}(\boldsymbol{F}_\text{1}),
\end{aligned}
\end{equation}
where $\boldsymbol{V}_\text{MES}$ denotes the input of the MES, $\boldsymbol{f}_\text{conv}(\cdot)=\boldsymbol{f}_\text{leakyrelu} (\boldsymbol{f}_\text{bn}(\boldsymbol{f}_\text{conv2d}(\cdot)))$ denotes the convolutional module, $\boldsymbol{f}_\text{conv2d}(\cdot)$ denotes the convolutional layer, $\boldsymbol{f}_\text{bn}(\cdot)$ denotes the batch normalization layer, and $\boldsymbol{f}_\text{leakyrelu}(\cdot)$ denotes the leaky ReLU activation function.
Then, we input $\boldsymbol{F}_2$ into the downsampling module $\boldsymbol{f}_\text{down}(\cdot)$ to obtain the multi-scale feature maps $\boldsymbol{F}_\text{3},\boldsymbol{F}_\text{4},\boldsymbol{D}_\text{1}$, which are defined by the following equations:
\begin{equation}
\label{eq:4}
\begin{aligned}
\boldsymbol{F}_\text{3} &= \boldsymbol{f}_\text{down}(\boldsymbol{F}_\text{2}), \\
\boldsymbol{F}_\text{4} &= \boldsymbol{f}_\text{down}(\boldsymbol{F}_\text{3}), \\
\boldsymbol{D}_\text{1} &= \boldsymbol{f}_\text{down}(\boldsymbol{F}_\text{4}), \\
\end{aligned}
\end{equation}
where $\boldsymbol{f}_\text{down}(\cdot)=\boldsymbol{f}_\text{conv}(\boldsymbol{f}_\text{conv}(\boldsymbol{f}_\text{conv}(\boldsymbol{f}_\text{avg-pool}(\cdot))))$ denotes the downsampling module, $\boldsymbol{f}_\text {avg-pool}(\cdot)$ denotes the average pooling layer (kernel size is 2, stride is 2, padding is 0).
And, we then decode the multi-scale features using the decoder module $\boldsymbol{f}_\text{spadeup}(\cdot)$ and simultaneously utilize motion confidence maps to modulate the parameters of the feature maps $\boldsymbol{D}_\text{2},\boldsymbol{D}_\text{3},\boldsymbol{D}_\text{4}$, which can be expressed as follows:
\begin{equation}
\label{eq:5}
\begin{aligned}
\boldsymbol{D}_\text{2} &= \boldsymbol{f}_\text{spadeup}(\boldsymbol{D}_\text{1},\boldsymbol{M},\boldsymbol{F}_\text{4}),\\
\boldsymbol{D}_\text{3} &= \boldsymbol{f}_\text{spadeup}(\boldsymbol{D}_\text{2},\boldsymbol{M},\boldsymbol{F}_\text{3}),\\
\boldsymbol{D}_\text{4} &= \boldsymbol{f}_\text{spadeup}(\boldsymbol{D}_\text{3},\boldsymbol{M},\boldsymbol{F}_\text{2}),\\
\end{aligned}
\end{equation}
where $\boldsymbol{M}(x,y) = e^{-\frac{t(x,y)-t_\text{max}}{\mathrm{\tau}}}\in[0,1]$ denotes the motion confidence map ($\mathrm{\tau}$ is a fixed-time constant).  
The value of $\boldsymbol{M}$ is $0$ in locations without event data. This method is inspired by \cite{lagorce2016hots}, but differs in the definition of $t_\text{max}$, which represents the maximum timestamp in the event stream.
Motion confidence maps record the motion information of the scene within a time window. The closer the time of motion to the current time, the higher the confidence value, otherwise the confidence value is lower. In order to increase the distinction between different times, the confidence value decays exponentially with time. Therefore, motion confidence maps can clearly distinguish the edge of the object and the texture within objects at the current time while preserving the past motion information. 
In more detail, $\boldsymbol{f}_\text{spadeup}(\cdot)$ is defined by the following equations:
\begin{equation}
\label{eq:6}
\begin{aligned}
\hat{\boldsymbol{D}}_{l-1} &= \boldsymbol{f}_\text{spade}(\boldsymbol{f}_\text{up}(\boldsymbol{D}_{l-1}),\boldsymbol{f}_\text{down}(\boldsymbol{M})),\\
\boldsymbol{D}_{l} &=\boldsymbol{f}_\text{conv}(\boldsymbol{f}_\text{conv}(\boldsymbol{F}_{6-l} \oplus \boldsymbol{f}_\text{conv}(\hat{\boldsymbol{D}}_{l-1}))),
\end{aligned}
\end{equation}
where $\hat{\boldsymbol{D}}_{l-1}$ denotes the result of modulating $\boldsymbol{D}_{l-1}$ with the function $\boldsymbol{f}_\text{spade}(\cdot)$, $\oplus$ denotes concatenation along the feature channel and $\odot$ denotes hadamard product. $\boldsymbol{f}_\text{up}(\cdot)$ denotes the bilinear upsampling layer (scale factor is 2), $\boldsymbol{f}_\text{down}(\cdot)$ denotes the nearest-neighbor downsampling, to adjust the resolution of $\boldsymbol{M}$ to match that of the upsampled $\boldsymbol{D}_{l-1}$. First, the upsampled $\boldsymbol{D}_{l-1}$ is modulated using $\boldsymbol{f}_\text{spade}(\cdot)$, and then concatenated with the encoder feature $\boldsymbol{F}_{6-l}$ along the channel dimension. 
Here, the function $\boldsymbol{f}_\text{spade}(\cdot)$ modulates the normalized and upsampled $\boldsymbol{D}_{l-1}$ using the scale and bias learned through downsampled $\boldsymbol{M}$ via a convolutional layer\cite{park2019semantic}. The structure of the core module $\boldsymbol{f}_{\text{spade}}(\cdot)$ within $\boldsymbol{f}_{\text{spadeup}}(\cdot)$ is shown in Fig. \ref{fig:3}. Its definition is as follows in detail: 
\begin{equation}
\label{eq:7}
\begin{aligned}
\boldsymbol{M}^\prime &= \boldsymbol{f}_\text{relu}(\boldsymbol{f}_\text{conv2d}(\boldsymbol{M}_{\downarrow})),\\
\boldsymbol{\gamma} &= 1+\boldsymbol{f}_\text{conv2d}(\boldsymbol{M}^\prime),\\
\boldsymbol{\beta} &= \boldsymbol{f}_\text{conv2d}(\boldsymbol{M}^\prime),\\
\hat{\boldsymbol{D}}_{l-1} &= \boldsymbol{f}_\text{bn-no-affine}(\boldsymbol{D}_{l-1\uparrow})\odot \boldsymbol{\gamma} + \boldsymbol{\beta},
\end{aligned}
\end{equation}
where $f_\text{bn-no-affine}(\cdot)$ is a batch normalization layer without learnable affine parameters, $\boldsymbol{D}_{l-1\uparrow}$ denote the upsampled $\boldsymbol{D}_{l-1}$ and $\boldsymbol{M}_{\downarrow}$ denote the downsampled $\boldsymbol{M}$.
Finally, a convolutional layer and softmax are used to output the weighted matrix $\boldsymbol{W}$ corresponding to the $\boldsymbol{V}_\text{MES}$, which is defined as follows:
\begin{equation}
\label{eq:8}
\boldsymbol{W} = \boldsymbol{f}_\text{softmax}(\boldsymbol{f}_\text{conv2d}(\boldsymbol{D}_4)),
\end{equation}
where $\boldsymbol{f}_\text{softmax}(\cdot)$ denotes the softmax layer. $\boldsymbol{E}_\text{EAA}\in\mathbb{R}^{\mathrm{H}\times\mathrm{W}}$ is the aggregated edge-modulated event frame generated by the EAA module after fusing motion confidence maps and the event frame, aggregated edge-modulated event frame specifically expressed as follows:
\begin{equation}
\label{eq:9}
\boldsymbol{E}_\text{EAA}(x,y)=\sum_{i=1}^\mathrm{L} \boldsymbol{W}(x,y,i)\odot \boldsymbol{V}_\text{MES}(x,y,i),
\end{equation}
where $\boldsymbol{V}_\text{MES}$  is defined in Eq. (\ref{eq:2}).
Specifically, we use a normalized weighting matrix $\boldsymbol{W}$ to perform weighted summation on ${V}_\text{MES}$ and generate the aggregated edge-modulated event frame, where $y$ and $x$ represent the pixel position, $i$ is the channel index of the ${V}_\text{MES}$, $L$ denotes the total number of channels. And each channel of the ${V}_\text{MES}$ is then weighted and summed using $\boldsymbol{W}$ to obtain the aggregated edge-modulated event frame.
In conclusion, by leveraging the clear object boundary and the texture within objects information provided by motion confidence maps, the EAA module combines multiple event accumulation frames $\boldsymbol{V}_\text{MES}$ from different time windows to generate the aggregated edge-modulated event frame $\boldsymbol{E}_\text{EAA}$ with improved clarity.

\subsection{Motion-Guided Attention Module (MGA Module) }
After generating left and right aggregated edge-modulated event frames, we use the MGA module to enhance the edge features of left and right aggregated edge-modulated event frames before constructing the cost volume.
Cost volume construction is a core component in stereo disparity estimation.
For deep learning-based approaches\cite{chang2018pyramid,zhang2019ga,khamis2018stereonet}, the left and right images are used as feature maps in the network to construct the cost volume.
Therefore, incorrect feature maps can significantly degrade the quality of the cost volume.
The pixel shifts in event frames can affect the formation of feature maps, thereby reducing the quality of cost volume construction.
Previous works overlook the preprocessing of feature maps that may experience pixel shifts before cost volume construction.
To address this issue, we propose a motion-guided attention (MGA) module. The implementation is detailed as follows.
Firstly, let $\boldsymbol{F}^{\text{f}}_{1, \text{l/r}}$, $\boldsymbol{F}^{\text{f}}_{2, \text{l/r}}$ and $\boldsymbol{F}^{\text{f}}_{3, \text{l/r}}$ denotes the multi-scale feature maps extracted from the left/right aggregated edge-modulated event frames $\boldsymbol{E}_\text{EAA}$ using the multi-scale feature extraction. These feature maps are subsequently utilized as inputs to the MGA module. And let $\boldsymbol{F}^{\text{m}}_{1, \text{l/r}}$, $\boldsymbol{F}^{\text{m}}_{2, \text{l/r}}$ and $\boldsymbol{F}^{\text{m}}_{3, \text{l/r}}$ be the multi-scale feature map multiplied by a motion confidence map $\boldsymbol{M}$, the following calculation is performed:
\begin{equation}
\label{eq:10}
\boldsymbol{F}_{l, \text{l/r}}^{\text{m}}=\boldsymbol{F}_{l, \text{l/r}}^{\text{f}}\odot\boldsymbol{M},\quad{l}=1,2,3,
\end{equation}
next, we combine the sets $\{\boldsymbol{F}_{l, \text{l/r}}^{\text{f}}\}^{3}_{l=1}$ and $\{\boldsymbol{F}_{l, \text{l/r}}^{\text{m}}\} ^{3}_{l=1}$ into a new set $\left\{\boldsymbol{x}_{l, \text{l/r}}\right\}_{l=1}^{6}$ and add positional and scale encoding to $\left\{\boldsymbol{x}_{l, \text{l/r}}\right\}_{l=1}^{6}$, we denote the resulting set as $\boldsymbol{X}$.
Then, we input the $\boldsymbol{X}$ into the deformable encoder, which is applied as follows:
\begin{equation}
\label{eq:11}
\boldsymbol{f}_\text{self-attn}(\boldsymbol{X})=\boldsymbol{f}_\text {LN}(\boldsymbol{f}_\text{deform-attn}(\boldsymbol{X})+\boldsymbol{X}),
\end{equation}
\begin{equation}
\label{eq:12}
\boldsymbol{Y}=\boldsymbol{f}_\text{LN}(\boldsymbol{f}_\text{FFN}(\boldsymbol{f}_\text{self-attn}(\boldsymbol{X}))),
\end{equation}
where $\boldsymbol{f}_\text{deform-attn}(\cdot)$ denotes deformable attention\cite{deformabledetr}, $\boldsymbol{f}_\text {LN}(\cdot)$ denotes layernorm layer and $\boldsymbol{f}_\text{FFN}(\cdot)$ denotes feed-forward network consisting of two MLP layers. Unlike global attention in \cite{Attention_is_all_you_need}, deformable attention only samples $\mathrm{K}$ points around the reference point to calculate the attention result. $\mathrm{K}$ reference points $s_{lk}$ are defined for each scale of $\boldsymbol{X}$. Subsequently, two linear layers $\boldsymbol{f}_\text{linear}(\cdot)$ are utilized to generate $\mathrm{K}$ sampling offsets $\Delta \boldsymbol{s}_{lk}$ and attention weights $\boldsymbol{A}_{lk}$. These offsets $\Delta \boldsymbol{s}_{lk}$ are then added to the coordinates of the reference points $\boldsymbol{s}_{lk}$, allowing the sampling of reference features $\boldsymbol{S}_{lk}$ from $\boldsymbol{X}$ through the utilization of bilinear interpolation $\boldsymbol{f}_\text{bilinear}(\cdot)$. 
Finally, the aggregation of $\boldsymbol{X}$ with the reference features $\boldsymbol{S}_{lk}$ is performed using the corresponding weights $\boldsymbol{A}_{lk}$, resulting in the derivation of the edge-enhanced feature map $\hat{\boldsymbol{X}}$. The $\boldsymbol{f}_\text{deform-attn}(\cdot)$ can be described by the following formulation:
\begin{equation}
\label{eq:13}
\boldsymbol{A}_{lk}=\boldsymbol{f}_\text{linear}(\boldsymbol{X}), \Delta \boldsymbol{s}_{lk}=\boldsymbol{f}_\text{linear}(\boldsymbol{X}),
\end{equation}
\begin{equation}
\label{eq:14}
\boldsymbol{S}_{l k}=\boldsymbol{f}_\text{bilinear}(\boldsymbol{X}, \boldsymbol{s}_{l k}+\Delta \boldsymbol{s}_{l k}),
\end{equation}
\begin{equation}
\label{eq:15}
\hat{\boldsymbol{X}}=\sum_{m=1}^{\mathrm{M}} \boldsymbol{W}_{m}( \sum_{l=1}^{\mathrm{L}} \sum_{k=1}^{\mathrm{K}} \boldsymbol{A}_{mlk} \cdot \boldsymbol{S}_{mlk}),
\end{equation}
where $\boldsymbol{W}_{m}$ denotes the learnable weights and $\mathrm{M}$, $\mathrm{L}$, $\mathrm{K}$ denote the number of attention heads, scales and reference points respectively. The feature maps extracted from $\boldsymbol{Y}$, specifically referring to the last three scales in terms of order, are denoted as $\{\boldsymbol{F}_{l, \text{l/r}}^{\text{MGA}}\} ^{3}_{l=1}$, corresponding to the order of $\{\boldsymbol{F}_{l, \text{l/r}}^{\text{m}}\} ^{3}_{l=1}$ at the input. These feature maps $\{\boldsymbol{F}_{l, \text{l/r}}^{\text{MGA}}\} ^{3}_{l=1}$ are used as input to the subsequent matching module.
At the same time, the first three scales are omitted in terms of order.
In conclusion, we utilize multi-scale deformable attention to compensate the multi-scale feature maps of event frames to the event frame feature maps multiplied by a motion confidence map. MGA module fuses information from motion confidence maps and event frames, complementing each other. This module enables the resulting feature maps used for cost volume construction to have more accurate edge structures and retain a significant portion of the geometric information.

\subsection{Disparity Estimation Network}
After the generation of multi-scale left and right feature maps $\{\boldsymbol{F}_{l, \text{l/r}}^{\text{MGA}}\} ^{3}_{l=1}$ is completed, we estimate the disparity by constructing and aggregating cost volumes and utilizing the WTA. The implementation is detailed as follows.
In our approach, we utilize the correlation method to construct a cost volume pyramid, which calculates the inner product of the feature vectors of left and right feature maps at each pixel to obtain the cost volume. The cost volume calculation can be formulated as follows:
\begin{equation}
\label{eq:16}
\begin{aligned}
\mathbf{C}^{\text {corr}}_{l}(d, x, y)=\frac{1}{\mathrm{C}}\left\langle\boldsymbol{F}^{\text{MGA}}_{l,\text{l}}(x,y), \boldsymbol{F}^\text{MGA}_{l,\text{r}}(x-d,y)\right\rangle,
\end{aligned}
\end{equation}
where $\langle\cdot, \cdot\rangle$ denotes the inner product between two feature maps, $\mathrm{C}$ represents the number of channels in a feature map, $l \in \{1,2,3\}$ denotes the scale index, and $\mathbf{C}^{\text {corr}}_{l}(d, x, y)$ represents the matching cost at position $(x,y)$ for a disparity of $d$. We use the intra-scale aggregation (ISA) module and cross-scale aggregation (CSA) module proposed in \cite{xu2020aanet} to construct the cost aggregation module. The ISA module performs deformable convolution on cost volumes of different scales while keeping the input and output tensor sizes the same. The CSA module utilizes equations Eq. (\ref{eq:17}) and Eq. (\ref{eq:18}) to perform cross-scale feature aggregation on the cost volumes output by the ISA module and the equations are as follows:
\begin{equation}
\label{eq:17}
\begin{aligned}
\hat{\mathbf{C}}_{l}=\sum_{k=1}^{3} \boldsymbol{f}_{k}\left(\tilde{\mathbf{C}}^{\text{corr}}_{l}\right), \quad l=1,2,3,
\end{aligned}
\end{equation}
where $\tilde{\mathbf{C}}^{\text{corr}}_{l}$ denotes the aggregated cost volume for the $k$-th scale $\mathbf{C}^{\text {corr}}_{l}$ after applying the ISA module, and $\hat{\mathbf{C}}_{l}$ denotes the aggregated cost volume after applying the CSA module. The structure of the $\boldsymbol{f}_{k}(\cdot)$ is as follows in detail:
\begin{equation}
\label{eq:18}
\begin{aligned}
\boldsymbol{f}_{k}(\tilde{\mathbf{C}}^{\text{corr}}_{l})=\left\{
\begin{array}{ll}
\tilde{\mathbf{C}}^{\text{corr}}_{l}    &\quad k=l\\
\boldsymbol{f}_{3 \times 3 \text{convs}}(\tilde{\mathbf{C}}^{\text{corr}}_{l})   &\quad k<l\\
\boldsymbol{f}_{1 \times 1 \text{conv}}(\boldsymbol{f}_{\text{up}}(\tilde{\mathbf{C}}^{\text{corr}}_{l})) &\quad k>l
\end{array}\right.,
\end{aligned}
\end{equation}
if $k<l$, $\boldsymbol{f}_{3\times3 \text{convs}}(\cdot)$ denotes the application of a $3\times3$ convolution layer repeated $(l-k)$ times with a stride of 2. If $k>l$,  $\boldsymbol{f}_\text{up}(\cdot)$ denotes the bilinear upsampling layer (scale factor is 2), $\boldsymbol{f}_{1\times1 \text{conv}}(\cdot)$ denotes $1\times1$ convolution layer. After aggregating the cost volume through the ISA and CSA modules, we utilize the WTA method to get multi-scale predicted disparity map through $\{\hat{\mathbf{C}}_{l}\}^{3}_{l=1}$. Finally, we utilize a refinement module\cite{refinement} to recover the multi-scale predicted disparity maps scale.

\subsection{Loss Function}

The underlying principle of stereo disparity estimation relies on the assumption that the left and right images demonstrate left-right consistency\cite{gosta2010accomplishments}. However, due to pixel shifts in event frames, the left-right consistency is compromised, leading to incomplete alignment between the two images.
Previous works overlook the supervision of left-right consistency in event representation within the network. To address this issue, we propose the left-right consistency census loss.
We use the smooth $L_{1}$ loss to supervise our network based on groundtruth, and introduce our proposed left-right consistency census loss on top of it to supervise simultaneously during network training. Our loss function can be divided into the following two parts.

\subsubsection{Left-Right Consistency Census Loss (census loss) }
we introduce a left-right consistency census loss function after the EAA module, which is formulated as: 
\begin{equation}
\label{eq:19}
\begin{aligned}
{L}_\text{census}=\sqrt{\left \|\boldsymbol{f}_\text{census}(\boldsymbol{I}_\text{left})-\boldsymbol{f}_\text{census}(\hat{\boldsymbol{I}}_\text{left})\right\| ^{2}+\epsilon^{2}}, 
\end{aligned}
\end{equation}
where $\boldsymbol{f}_\text{census}(\cdot)$ refers to the census transform, $\epsilon$ is a diminutive constant. Firstly, $\boldsymbol{\hat{I}}_\text{left}$ is obtained from $\boldsymbol{I}_\text{right}$ through a warp operation, and the warp operation is as follows:
\begin{equation}
\label{eq:20}
\begin{aligned}
\hat{\boldsymbol{I}}_\text{left}(x,y) = \boldsymbol{I}_\text{right}\left(x - d(x,y), y\right),
\end{aligned}
\end{equation}
where $\hat{\boldsymbol{I}}_\text{left}$ denotes the left event frame warped from the right event frame using the ground truth disparity $d$. Finally, $\boldsymbol{f}_\text{census}(\cdot)$ is defined as follows:
\begin{equation}
\label{eq:21}
\begin{aligned}
&\boldsymbol{f}_\text{census}(\boldsymbol{I})(x,y)=\\
&\sum_{i=-k}^k{\sum_{j=-k}^k{\left[\text{sgn}\left(\boldsymbol{I}(x+i,y+j)-\boldsymbol{I}(x,y)\right)\right]\times2^{|i|+|j|}}},
\end{aligned}
\end{equation}
where $\boldsymbol{I}$ denotes the original grayscale values, $k$ denotes the window size of the Census algorithm, and $\text{sgn}$ denotes the sign function. 
We compute the difference in geometric structure between the left event frame transformed by the census transform, and the left event frame warped from the right event frame, using the charbonnier loss\cite{sun2010secrets}, and use this difference as the loss function for the stereo event frame. Due to the potentially different number of events between the two cameras, we utilize the census transform on the left and right images, which only considers the relative positions of pixel intensities, instead of the traditional photometric consistency loss. The census transform maps each pixel of a grayscale image to an 8-bit binary code that encodes whether the pixel's intensity value is greater than that of its surrounding pixels. This non-parametric transform depends only on the relative ordering of intensity values and not their actual values.
In conclusion, left-right consistency census loss can further supervise the left-right consistency of aggregated edge-modulated event frames generated by the EAA module.
\subsubsection{Total Loss Function}
We utilize both the smooth $L_{1}$ loss function\cite{girshick2015fast} and the left-right consistency census loss to construct the total loss function. The smooth $L_{1}$ loss function is computed as follows:
\begin{equation}
\label{eq:22}
\begin{aligned}
L_\text{smooth-L1}(x) = \begin{cases}
0.5x^{2} , & \text{ if } |x|<1, \\
|x| - 0.5, & \text{ otherwise}, 
\end{cases} 
\end{aligned}
\end{equation}
and, the total loss function is computed as follows:
\begin{equation}
\label{eq:23}
\begin{aligned}
L = \sum_{l = 1}^{5} \lambda_{l}  \cdot L_\text{smooth-L1}(d_{l} - \hat{d}_{l}) + \lambda \cdot {L}_\text{census}(\boldsymbol{I}_\text{left}, \boldsymbol{I}_\text{right}, d_{l}),
\end{aligned}
\end{equation}
where $d$ denotes the ground truth disparity, $\hat{d}$ denotes the predicted disparity, $\lambda$ denotes the weight of the left-right consistency census loss, and $\lambda_{l}$ represents the weight of the smooth $L_{1}$ loss for each scale. The $5$ scales refer to the original size, $1/2$, $1/3$, $1/6$ and $1/12$, where the last two scales are output by the refinement module.

\section{Experiments}
\label{sec:experiments}
In this section, we evaluate our method on DSEC dataset\cite{DSEC} and conduct qualitative and quantitative comparison experiments with previous state-of-the-art methods, demonstrating the effectiveness of our method. Subsequently, by performing ablation studies on the sub-datasets we partitioned, utilizing the modules and loss functions introduced in Section \ref{sec:methodology}, we validated the advantages of each individual module.

\subsection{Experimental Setup}
\subsubsection{Dataset Selection}
\label{Dataset Selection}
In order to compare our method with the state-of-the-art (SOTA) methods, we evaluate it on the DSEC disparity benchmark\cite{DSEC}. We follow the original training and testing set division of the DSEC dataset for our experiments. 
To conduct ablation experiments, we randomly select 2/3 of the total DSEC training set as the training set and the remaining 1/3 as the DSEC validation set.
For the ablation study focusing on night scenes, we select all DSEC night scene sequences from the validation set for experimentation.
\subsubsection{Parameters in Training}
We implement the proposed method using the PyTorch framework. To train the proposed method, we initialize the network weights randomly and perform end-to-end training using four RTX3090 GPUs. We use the Adam optimizer with beta values of (0.9, 0.999). To train the network, we use a batch size of 16 and train it on the entire DSEC training set for 100 epochs. We linearly increase the learning rate for the first 3 epochs and then anneal it using a cosine schedule with a weight decay of 1e-4.
\subsubsection{Evaluation Metrics}
we reference prior works \cite{tulyakov2019learning,mostafavi2021event,nam2022stereo,zhang2022discrete} and utilize evaluation metrics such as Root Mean Squared Error (RMSE), Mean Absolute Error (MAE), 1-Pixel Error (1PE) and 2-Pixel Error (2PE). The calculation equations for RMSE and MAE are as follows:
\begin{equation}
\label{eq:24}
\begin{aligned}
\text{RMSE}=\sqrt{\frac{1}{\mathrm{N}}\sum_{i=1}^\mathrm{N}\left(d(i)-\hat{d}(i)\right)^{2}},
\end{aligned}
\end{equation}

\begin{equation}
\label{eq:25}
\begin{aligned}
\text{MAE}=\frac{1}{\mathrm{N}} \sum_{i=1}^\mathrm{N}\left|d(i)-\hat{d}(i)\right|,
\end{aligned}
\end{equation}
where $\mathrm{N}$ denotes the number of effective pixels. The percentage of pixels with a disparity error greater than $\delta$ over the total number of pixels $\mathrm{N}$ is referred to as $\delta$-pixel error ($\delta$PE), where $\delta$ is 1 for 1PE and 2 for 2PE, respectively.

\begin{figure*}[htbp]
\vspace{-0.0em}
\centering
\subfigure[]{\label{fig:4_1}\includegraphics[width=0.197\textwidth]{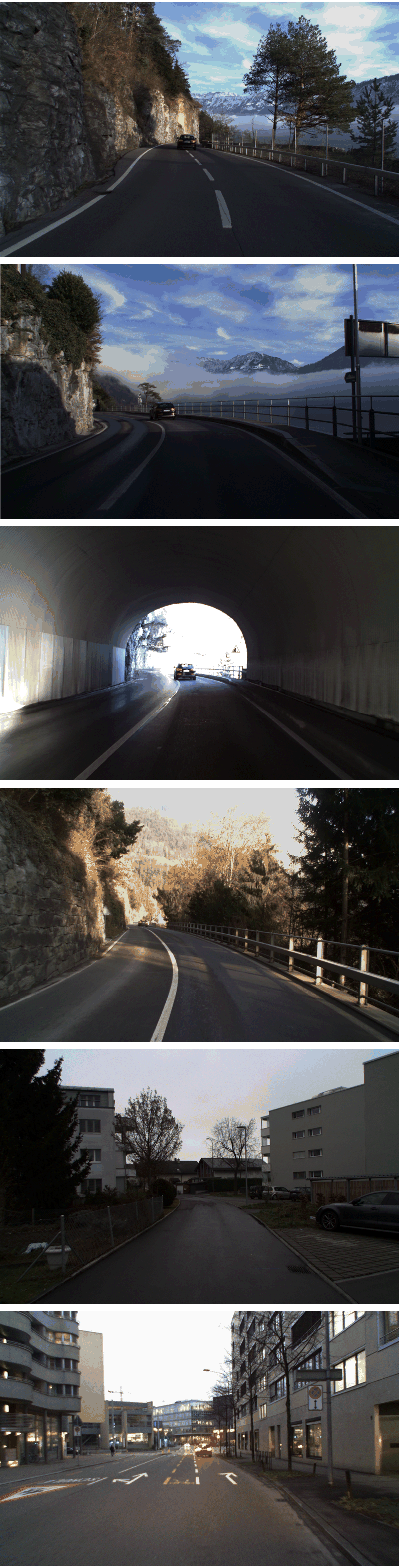}}
\hspace{-0.5em}
\subfigure[]{\label{fig:4_2}\includegraphics[width=0.197\textwidth]{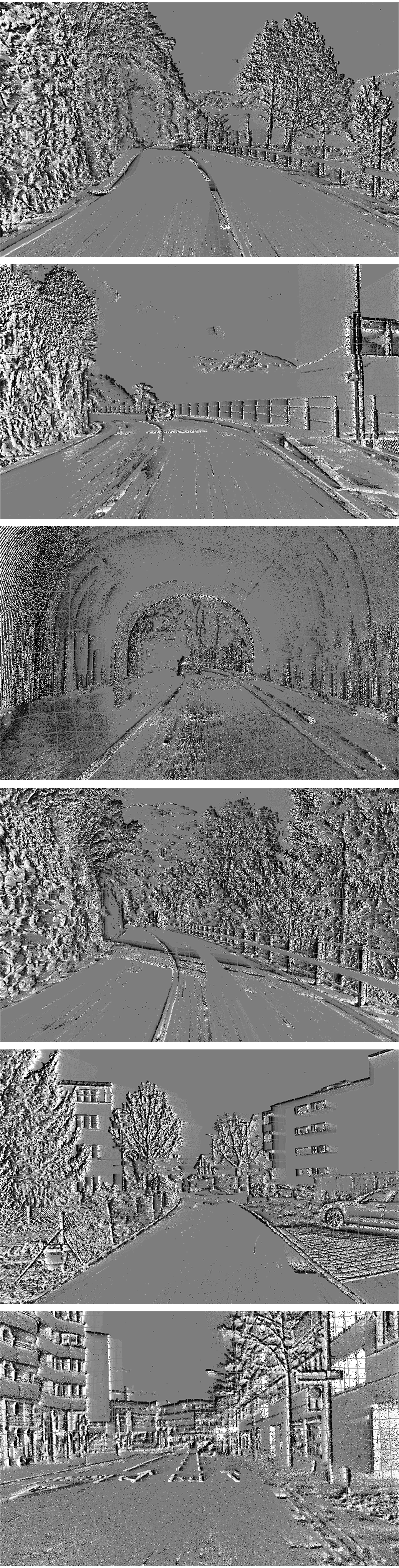}}
\hspace{-0.5em}
\subfigure[]{\label{fig:4_3}\includegraphics[width=0.197\textwidth]{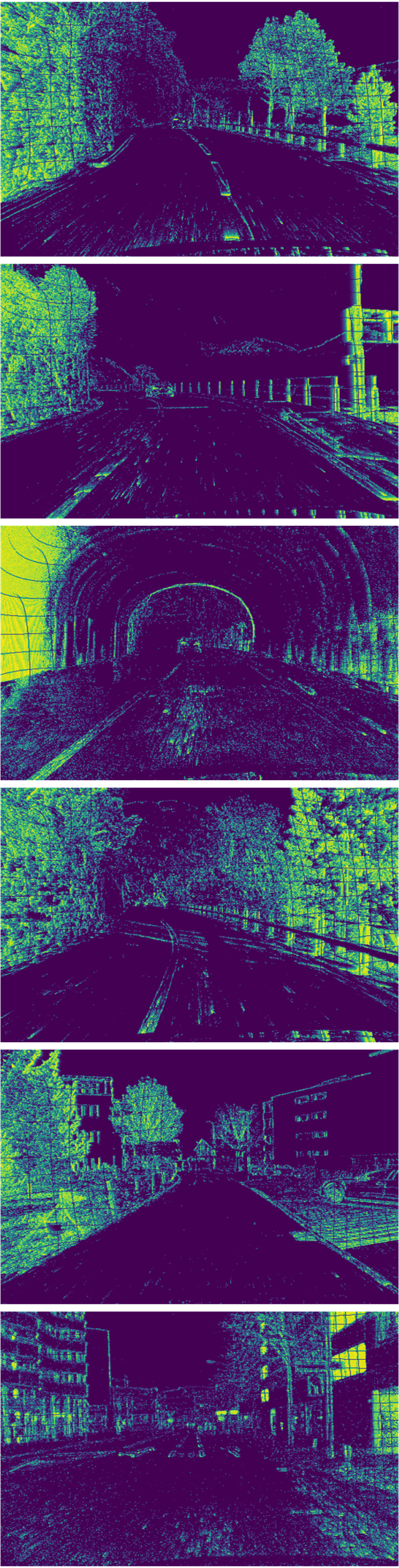}}
\hspace{-0.5em}
\subfigure[]{\label{fig:4_4}\includegraphics[width=0.197\textwidth]{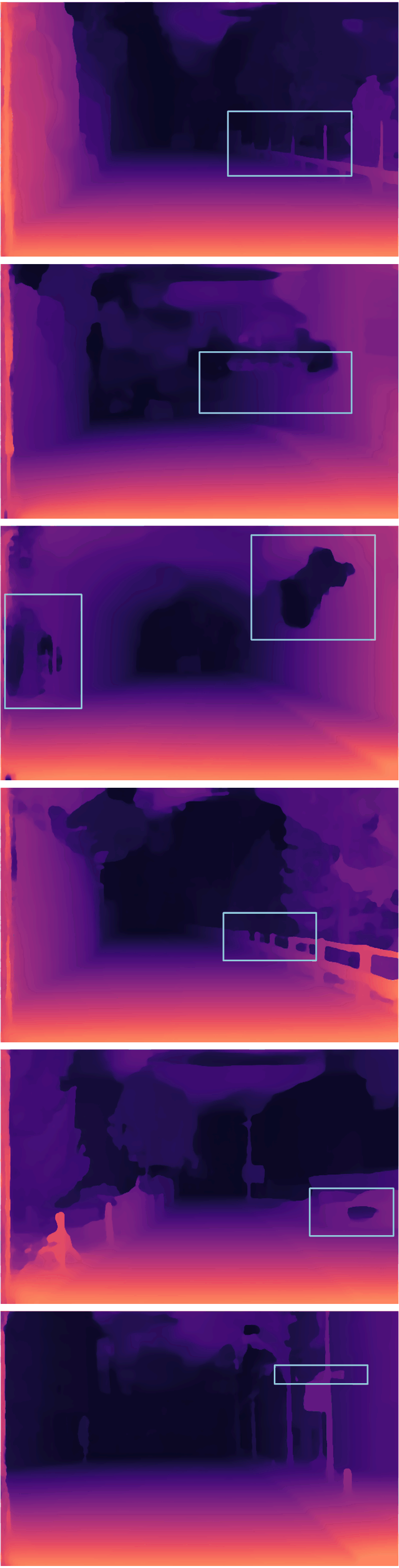}}
\hspace{-0.5em}
\subfigure[]{\label{fig:4_5}\includegraphics[width=0.197\textwidth]{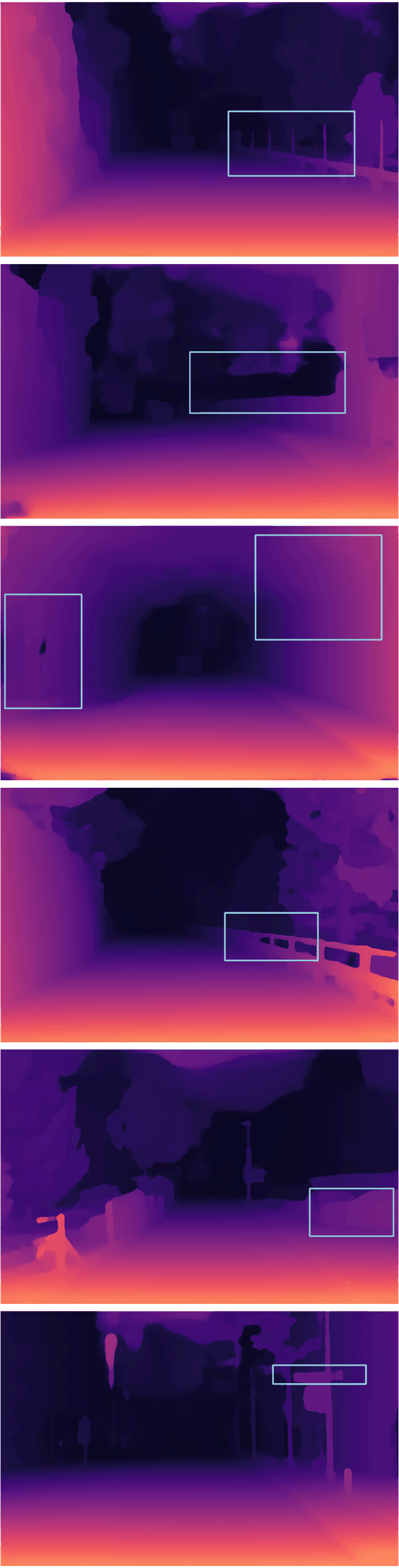}}
\vspace{-0.5em}
\caption{Qualitative evaluation of our best model compared with SOTA models on sequences from the DSEC\cite{DSEC} dataset. These scenarios comprise three locations: Interlaken, Thun and Zurich City. (a) Images (only for visualization), (b) Aggregated edge-modulated event frames, (c) Motion confidence map (the brighter areas have higher confidence levels), (d) Disparity maps of Concentration Net\cite{nam2022stereo}, (e) Disparity maps of EV-MGDispNet (ours).}
\label{fig:4}
\vspace{-6.0em}
\end{figure*}

\begin{table*}[htbp]
\vspace{7.0em}
\begin{center}
\caption{Comparison of our method with SOTA disparity estimation methods\cite{tulyakov2019learning,mostafavi2021event,nam2022stereo,zhang2022discrete} on the DSEC\cite{DSEC} dataset. Best in bold, runner-up underlined, lower values denote better performance.}
\label{tab:1}
\setlength{\tabcolsep}{0.020\linewidth}
\begin{tabular}{c|c|c|c|c|c|c|c|c}
\hline
\multirow{2}{*}{Methods}
&\multicolumn{4}{c|}{MAE$\downarrow$}
&\multicolumn{4}{c}{RMSE$\downarrow$}\\
\cline{2-9} 
&Inter  &Thun   &City   &All
&Inter  &Thun   &City   &All\\
\hline
DDES\cite{tulyakov2019learning}
&0.573      &0.632      &0.564      &0.576
&1.364      &1.634      &1.332      &1.381\\
\hline
E-Stereo\cite{mostafavi2021event}
&---    &---    &---    &0.529
&---    &---    &---    &\underline{1.222}\\
\hline
Concentration Net\cite{nam2022stereo}
&\underline{0.514}  &0.553  &0.517  &\underline{0.519}
&\underline{1.207}  &1.483              &1.192  &1.231\\
\hline
DTC-PDS\cite{zhang2022discrete}
&{0.536}
&{\underline{0.55}}
&{\textbf{0.512}}
&{0.527}
&{1.299}
&{\underline{1.444}}
&{\underline{1.181}}
&{1.264}\\
\hline
EV-MGDispNet (Ours)
&\textbf{0.509} &\textbf{0.540} &\textbf{0.512} &\textbf{0.514}
&\textbf{1.161} &\textbf{1.400} &\textbf{1.158}     &\textbf{1.187}\\
\hline
\hline
\multirow{2}{*}{Methods}
&\multicolumn{4}{c|}{1PE$\downarrow$}
&\multicolumn{4}{c}{2PE$\downarrow$}\\
\cline{2-9} 
&Inter  &Thun   &City   &All
&Inter  &Thun   &City   &All\\
\hline
DDES\cite{tulyakov2019learning}
&10.671     &10.854     &11.184     &10.915
&3.125      &3.225      &2.593      &2.905\\
\hline
E-Stereo\cite{mostafavi2021event}
&---    &---    &---    &9.958
&---    &---    &---    &2.645\\
\hline
Concentration Net\cite{nam2022stereo}
&\underline{9.377}  &\textbf{8.506} &{\underline{10.075}}   &{\underline{9.583}}
&2.818  &2.924          &2.335                  &2.620\\
\hline
{DTC-PDS\cite{zhang2022discrete}}
&{9.554}
&{8.905}
&{\textbf{9.638}}
&{\textbf{9.517}}
&{\underline{2.654}}
&{\textbf{2.474}}
&{\textbf{2.015}}
&{\textbf{2.356}}\\
\hline
EV-MGDispNet (Ours)
&\textbf{9.220}  &\underline{8.750}       &10.272 &9.625
&\textbf{2.647}     &\underline{2.833}  &\underline{2.289}  &\underline{2.512}\\
\hline
\end{tabular}
\vspace{-1.0em}
\end{center}
\end{table*}

\begin{table*}[htbp]
\vspace{0.0em}
\begin{center}
\caption{Module ablation. Sequentially adding MGA module, EAA module and left-right consistency census loss improves the model's accuracy with each module and method. Best in bold, runner-up underlined, lower values denote better performance.}
\label{tab:2}
\setlength{\tabcolsep}{0.005\linewidth}
\begin{tabular}{c|cccc|cccc|cccc|cccc}
\hline
\multirow{2}{*}{Ablation Studies}
&\multicolumn{4}{c|}{MAE$\downarrow$}
&\multicolumn{4}{c|}{RMSE$\downarrow$}
&\multicolumn{4}{c|}{1PE$\downarrow$}
&\multicolumn{4}{c}{2PE$\downarrow$}\\
\cline{2-17}
&\multicolumn{1}{c|}{Inter}
&\multicolumn{1}{c|}{Thun}
&\multicolumn{1}{c|}{City}
&All
&\multicolumn{1}{c|}{Inter}
&\multicolumn{1}{c|}{Thun}
&\multicolumn{1}{c|}{City}
&All
&\multicolumn{1}{c|}{Inter}
&\multicolumn{1}{c|}{Thun}
&\multicolumn{1}{c|}{City}
&All
&\multicolumn{1}{c|}{Inter}
&\multicolumn{1}{c|}{Thun}
&\multicolumn{1}{c|}{City}
&All\\
\hline
\multicolumn{1}{l|}{EV-MGDispNet (Ours)}
&\multicolumn{1}{c|}{\textbf{0.617}}
&\multicolumn{1}{c|}{\textbf{0.796}}
&\multicolumn{1}{c|}{\textbf{0.603}}
&\textbf{0.612}
&\multicolumn{1}{c|}{\textbf{1.486}}
&\multicolumn{1}{c|}{\textbf{1.954}}
&\multicolumn{1}{c|}{\textbf{1.388}}
&\textbf{1.432}
&\multicolumn{1}{c|}{\textbf{12.097}}
&\multicolumn{1}{c|}{\textbf{18.072}}
&\multicolumn{1}{c|}{\textbf{12.565}}
&\textbf{12.559}
&\multicolumn{1}{c|}{\textbf{3.330}}
&\multicolumn{1}{c|}{\textbf{4.989}}
&\multicolumn{1}{c|}{\textbf{2.947}}
&\textbf{3.114}\\
\hline
\multicolumn{1}{l|}{\quad w/o census loss}
& \multicolumn{1}{c|}{\underline{0.634}}
& \multicolumn{1}{c|}{\underline{0.798}}
& \multicolumn{1}{c|}{\underline{0.609}}
& \underline{0.622}
& \multicolumn{1}{c|}{\underline{1.527}}
& \multicolumn{1}{c|}{\underline{1.967}}
& \multicolumn{1}{c|}{\underline{1.416}}
& \underline{1.463}
& \multicolumn{1}{c|}{\underline{12.343}}
& \multicolumn{1}{c|}{\underline{18.129}}
& \multicolumn{1}{c|}{\underline{12.576}}
& \underline{12.642}
& \multicolumn{1}{c|}{\underline{3.468}}
& \multicolumn{1}{c|}{\underline{5.082}}
& \multicolumn{1}{c|}{\underline{3.031}}
& \underline{3.214}\\
\hline
\multicolumn{1}{l|}{\quad w/o EAA, census loss}
& \multicolumn{1}{c|}{0.651}
& \multicolumn{1}{c|}{0.818}
& \multicolumn{1}{c|}{0.613}
& 0.629
& \multicolumn{1}{c|}{1.552}
& \multicolumn{1}{c|}{2.017}
& \multicolumn{1}{c|}{1.432}
& 1.482
& \multicolumn{1}{c|}{13.090}
& \multicolumn{1}{c|}{18.553}
& \multicolumn{1}{c|}{12.654}
& 12.932
& \multicolumn{1}{c|}{3.774}
& \multicolumn{1}{c|}{5.256}
& \multicolumn{1}{c|}{3.047}
& 3.322\\
\hline
\multicolumn{1}{l|}{\quad w/o MGA, EAA, census loss}
& \multicolumn{1}{c|}{0.679}
& \multicolumn{1}{c|}{0.840}
& \multicolumn{1}{c|}{0.622}
& 0.644
& \multicolumn{1}{c|}{1.658}
& \multicolumn{1}{c|}{2.106}
& \multicolumn{1}{c|}{1.434}
& 1.518
& \multicolumn{1}{c|}{13.261}
& \multicolumn{1}{c|}{18.755}
& \multicolumn{1}{c|}{13.088}
& 13.280
& \multicolumn{1}{c|}{4.116}
& \multicolumn{1}{c|}{5.307}
& \multicolumn{1}{c|}{3.126}
& 3.480\\
\hline
\end{tabular}
\end{center}
\end{table*}

\begin{table}[htbp]
\vspace{-0.0em}
\begin{center}
\caption{Night scene ablation. In the night scene sequence, sequentially adding MGA module, EAA module and left-right consistency census loss improve the model's accuracy with each module and method. Best in bold, runner-up underlined, lower values denote better performance.}
\label{tab:3}
\setlength{\tabcolsep}{0.01\linewidth}
\begin{tabular}{l|c|c|c|c}
\hline
\multicolumn{1}{c|}{Night Ablation}
&1PE$\downarrow$
&2PE$\downarrow$
&MAE$\downarrow$
&RMSE$\downarrow$\\
\hline
EV-MGDispNet (Ours) 
&\textbf{14.887}    &\textbf{4.207} &\textbf{0.657} &\textbf{1.487}\\
\hline
\quad w/o census loss
&\underline{14.904} &\underline{4.241}  &\underline{0.660}  &\underline{1.493}\\
\hline
\begin{tabular}[c]{@{}l@{}}
\quad w/o EAA, census loss
\end{tabular}
&15.186 &4.299  &0.666  &1.500\\
\hline
\begin{tabular}[c]{@{}l@{}}
\quad w/o MGA, EAA, census loss
\end{tabular}
&15.267 &4.414  &0.683  &1.561\\
\hline
\end{tabular}
\vspace{-2.0em}
\end{center}
\end{table}

\subsection{Comparison against SOTA Methods}
\subsubsection{Quantitative Analysis}
We compare our EV-MGDispNet with state-of-the-art algorithms (DDES\cite{tulyakov2019learning}, E-Stereo\cite{mostafavi2021event}, Concentration Net\cite{nam2022stereo}, DTC-PDS\cite{zhang2022discrete}) on 1PE, 2PE, MAE and RMSE metrics using the DSEC disparity benchmark dataset. All test data are evaluated on the server without disclosing the ground truth, making the comparison results reliable. As shown in Table \ref{tab:1}, we achieve the best performance in terms of MAE and RMSE metrics and reach a sub-optimal result in 2PE, slightly worse only in 1PE. Specifically, we compare the performance of our algorithm on three different scenes, Inter, Thun and City, as well as the overall performance on all test samples, denoted as ``all" against the earliest work in the field, DDES, as shown in Table \ref{tab:1}. Our four performance metrics, namely 1PE, 2PE, MAE and RMSE, significantly outperform the DDES algorithm across all scenes. In comparison with E-Stereo, although it achieves the lowest RMSE among all SOTA algorithms except for ours, our algorithm still outperforms it. When compared to Concentration Net, we achieve comparable results in 1PE, but lower 2PE, MAE and RMSE. In contrast, our algorithm showes better performance in terms of MAE and RMSE when compared with the DTC-PDS, and competitive results in 1PE and 2PE. Importantly, we achieve the best MAE, which is the main metric in DSEC.

The existing works \cite{zhang2022discrete, nam2022stereo, mostafavi2021event} mainly rely on components from the field of image-based stereo disparity estimation to perform basic functionalities such as cost volume computation, volume aggregation and disparity estimation. For this process, accurate volume calculation is essential for achieving precise disparity estimation. As shown in Table \ref{tab:2}, Table \ref{tab:3} and Table \ref{tab:4}, our EAA module integrates the motion confidence maps during the generation of aggregated edge-modulated event frames, which helps in creating event representations with more accurate boundaries. Additionally, as shown in Table \ref{tab:4}, our left-right consistency census loss imposes extra constraints on the consistency of left and right aggregated edge-modulated event frames after generating the aggregated edge-modulated event frames. Both of these aspects improve the left-right consistency of the generated left and right event representations before volume calculation, and the left-right consistency of the stereo task is fundamental for achieving high-precision stereo disparity estimation. Previous works \cite{nam2022stereo, mostafavi2021event} generate event representation through neural networks, but the generated event representation suffers from pixel shifts. Although the generated event representation can partially alleviate this problem, they cannot completely solve it. As shown in Table \ref{tab:2} and Table \ref{tab:3}, we improve the generation of event frames and employ the MGA module to mitigate the impact of these issues to some extent by incorporating motion confidence maps. Table \ref{tab:2}, Table \ref{tab:3} and Table \ref{tab:4} demonstrate the effectiveness of these improvements are crucial for enhancing performance. Additionally, Table \ref{tab:1} demonstrates that using these methods together can effectively improve the disparity estimation accuracy, bringing it to the state-of-the-art level.

It is worth noting that we do not use knowledge distillation and only utilize past events without incorporating future events, which has been shown in the ablation experiment of \cite{nam2022stereo} to improve the performance of the network proposed in \cite{nam2022stereo}. And we explicitly fuse information from event frames and motion confidence maps.

\subsubsection{Qualitative Analysis}
In Fig. \ref{fig:4}, we present a qualitative comparison of our method with \cite{nam2022stereo} in various scenarios. 
Our model exhibits superior qualitative estimation results compared to the SOTA model \cite{nam2022stereo}. 

\textbf{Regarding fine structures}, our proposed EAA module, MGA module and the left-right consistency census loss contribute to the improved reconstruction of fine structures, such as railings, road signs and obstacles, in the disparity estimated by our model.
We believe that the improved accuracy of fine structures may be attributed to the integration of motion information in the EAA module, which allows the generated aggregated edge-modulated event frame to possess more accurate edges and contours. 
Additionally, the edge-enhanced feature maps obtained through the MGA module exhibit 
features with reduced pixel shifts in terms of scene edges, texture within objects and object contours, owing to the guidance provided by motion confidence maps. This guidance has a more direct impact on improving the disparity estimation results at fine structures.
Finally, the left-right consistency census loss serves as a direct supervision for the generated results of the EAA module within the network. It enforces the generated aggregated edge-modulated event frame to adhere to the principle of left-right consistency in stereo cameras. By directly penalizing the EAA module's outputs that do not meet the left-right consistency criteria before feature extraction in the MGA module, this loss mitigates the impact of the left-right consistency loss on stereo disparity estimation results.

\textbf{Regarding planar structures}, our model exhibits fewer holes and smoother disparity estimation results in plane such as tunnels, walls and vehicle surfaces, thanks to our proposed MGA module. We believe that this improvement may be attributed to the MGA module's ability to extract multi-scale features from the aggregated edge-modulated event frame and perform feature fusion and complementarity using the multi-scale deformable attention mechanism. By incorporating the original multi-scale aggregated edge-modulated event frame features and the features multiplied by motion confidence maps, the MGA module effectively utilizes geometric information while preserving scene edges and object contours accurately. As a result, the MGA module reduces the hole effect in plane and further enhances the restoration of fine structures.

\begin{figure*}[htbp]
\centering
\subfigure[]{\label{fig:5_1}\includegraphics[width=0.1669\textwidth]{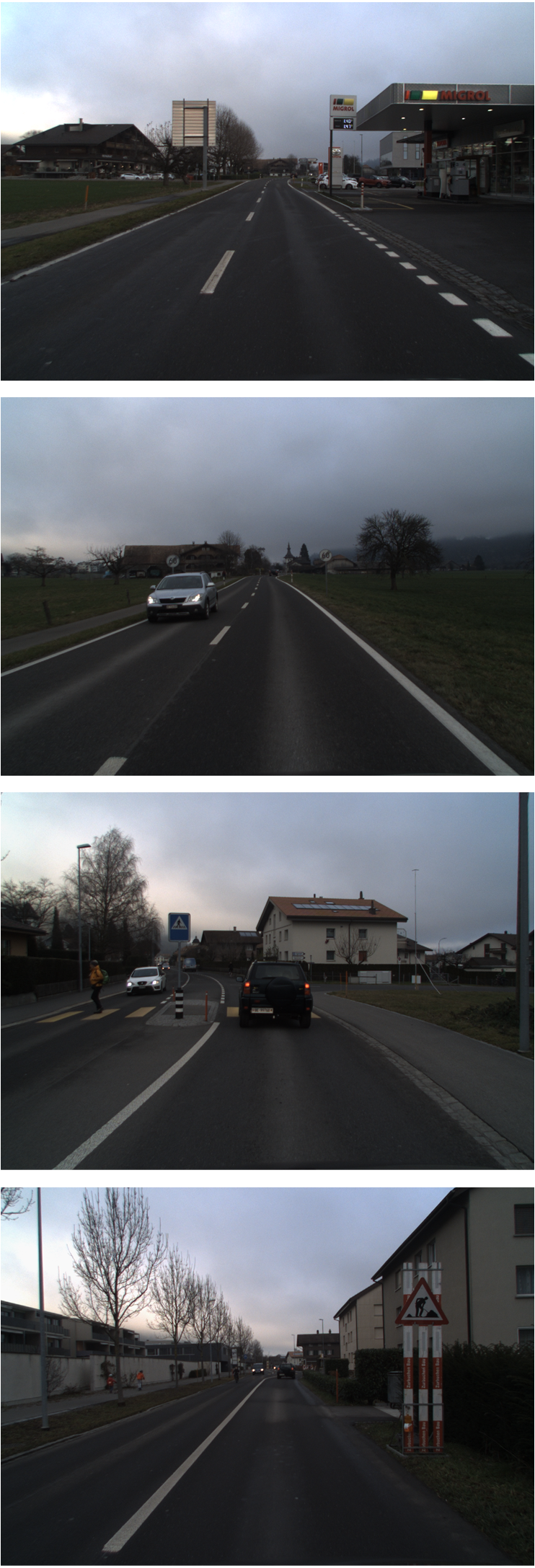}}
\hspace{-0.5em}
\subfigure[]{\label{fig:5_2}\includegraphics[width=0.2051\textwidth]{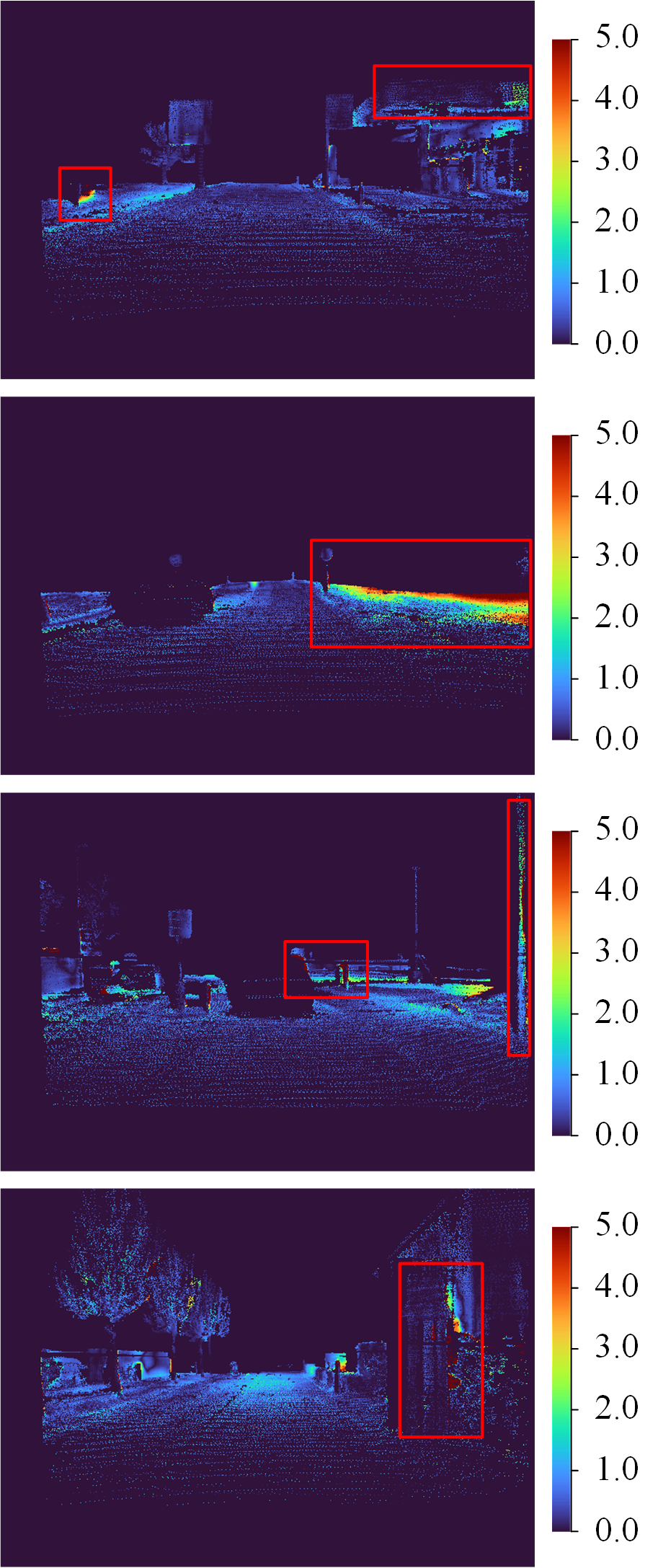}}
\hspace{-0.5em}
\subfigure[]{\label{fig:5_3}\includegraphics[width=0.2051\textwidth]{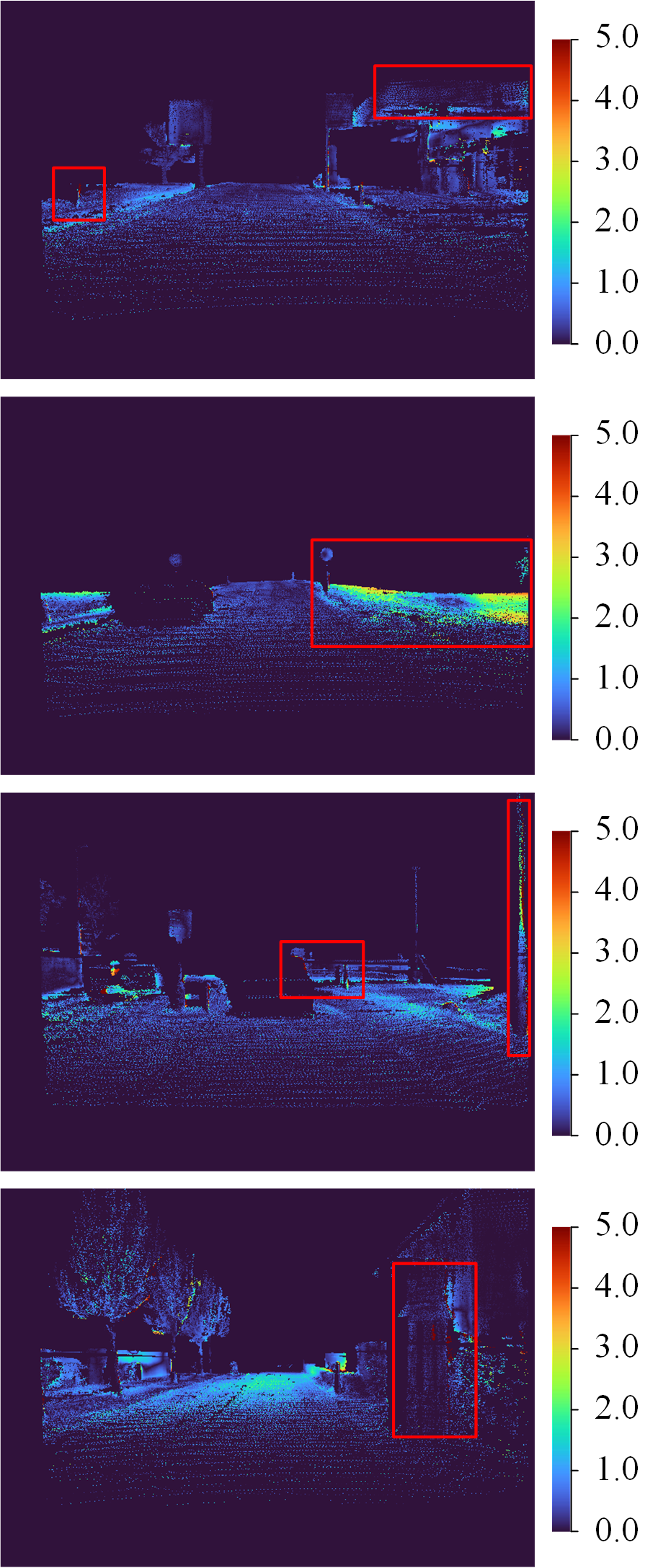}}
\hspace{-0.5em}
\subfigure[]{\label{fig:5_4}\includegraphics[width=0.2051\textwidth]{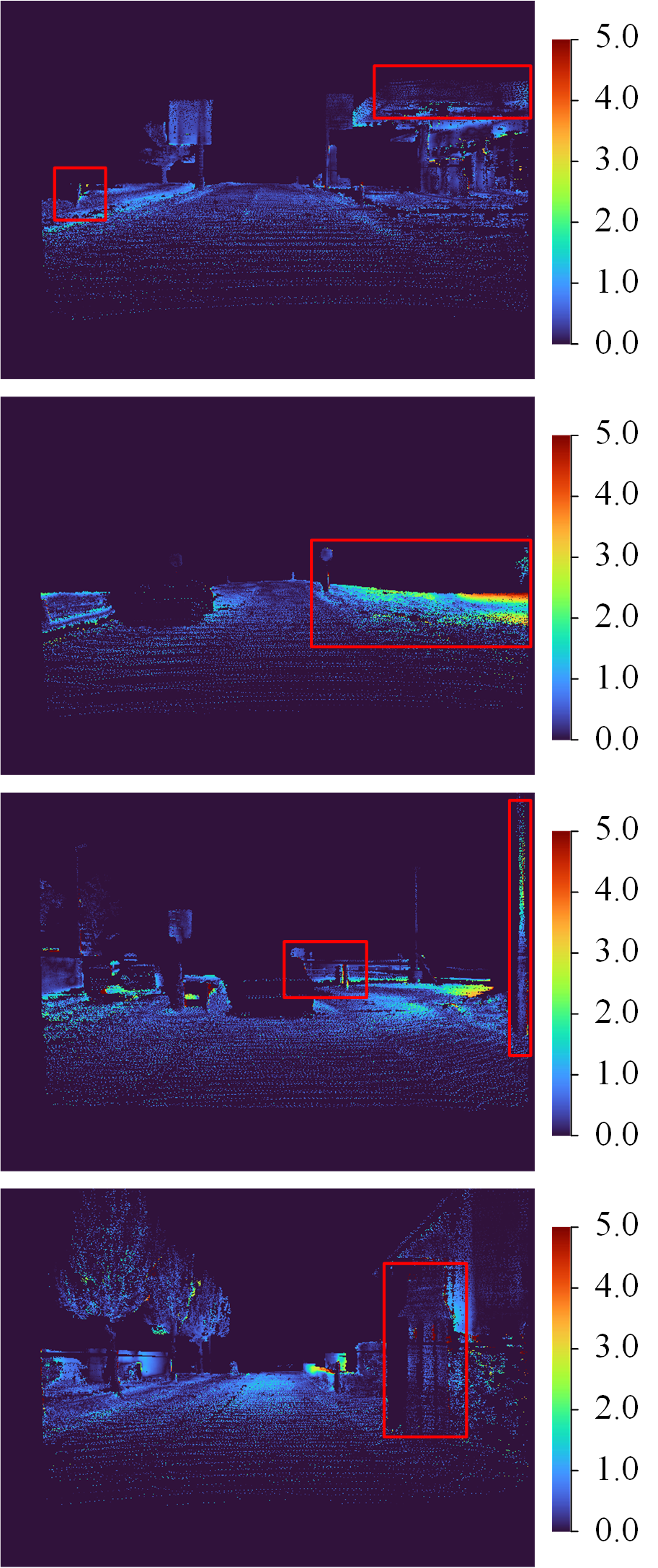}}
\hspace{-0.5em}
\subfigure[]{\label{fig:5_5}\includegraphics[width=0.2051\textwidth]{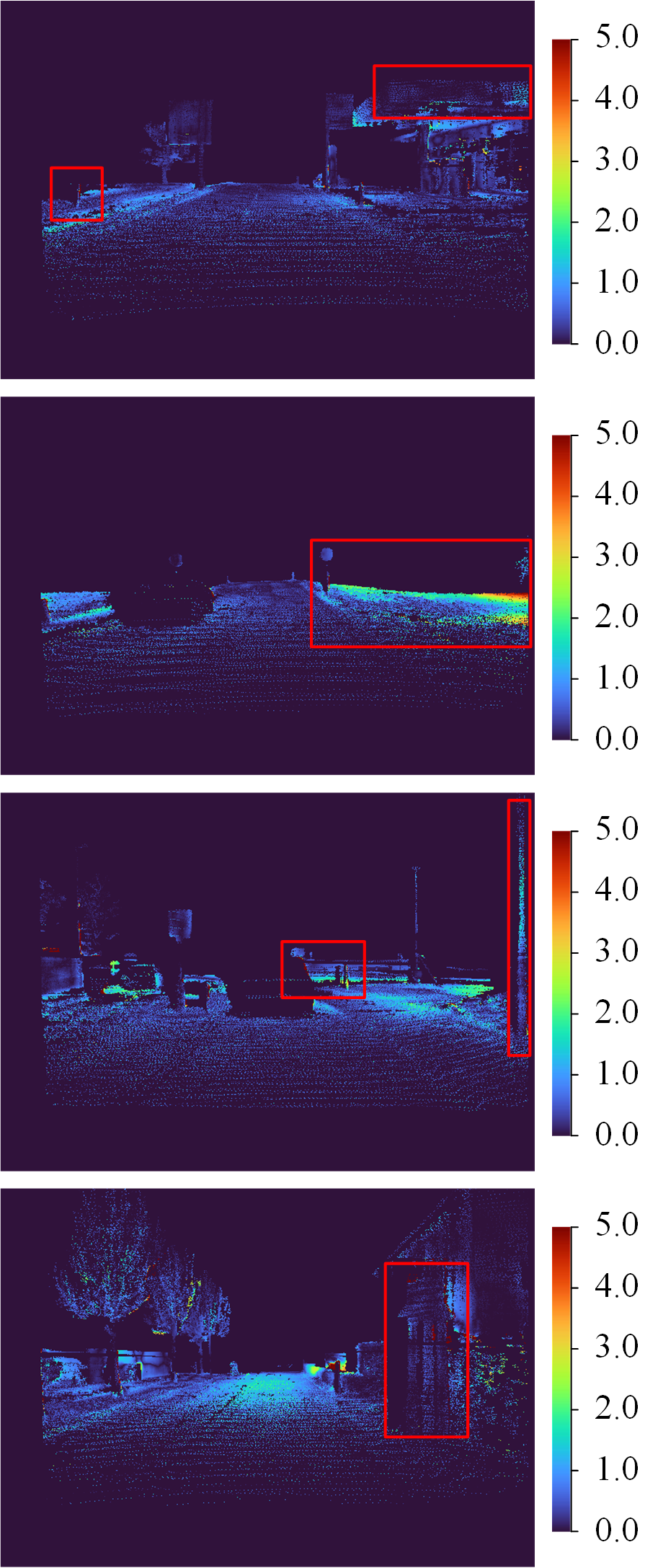}}
\vspace{-0.5em}
\caption{Module qualitative ablation. Sequentially adding MGA module, EAA module and left-right consistency census loss leads to a reduction in disparity estimation errors pertaining to scene contours and intricate structural elements. (a) Images (only for visualization) (b) w/o MGA module, EAA module, census loss, (c) w/o EAA module, census loss, (d) w/o cenloss, (e) EV-MGDispNet (Ours). }
\label{fig:5}
\vspace{-1em}
\end{figure*}

\subsection{Ablation Studies}

As shown in Table \ref{tab:2} and Table \ref{tab:3}, we conduct a series of ablation experiments in validation set and night scenes, progressive exclusion of EAA module, MGA module and left-right consistency census loss from EV-MGDispNet. Importantly, it should be noted that as each module is progressively excluded, the model's performance exhibits a persistent degradation. In Table \ref{tab:4}, we employ two fundamental metrics, namely Peak Signal-to-Noise Ratio (PSNR) and Structural Similarity (SSIM)\cite{wang2004image}, to evaluate the efficiency of the different configurations which sequentially remove the left-right consistency census loss and EAA modules from EV-MGDispNet. The ablation dataset is partitioned following the methodology outlined in Section \ref{Dataset Selection}. 

The calculation equations for PSNR and SSIM are as follows:
\begin{equation}
\label{eq:26}
\begin{aligned}
\text{PSNR}=10 \cdot \log _{10}\left(\frac{\text{MAX}_{I}^{2}}{\text{MSE}}\right),
\end{aligned}
\end{equation}
\begin{equation}
\label{eq:27}
\begin{aligned}
\operatorname{SSIM}(x, y)=\frac{\left(2 \mu_{x} \mu_{y}+\mathrm{C}_{1}\right)\left(2 \sigma_{x y}+\mathrm{C}_{2}\right)}{\left(\mu_{x}^{2}+\mu_{y}^{2}+\mathrm{C}_{1}\right)\left(\sigma_{x}^{2}+\sigma_{y}^{2}+\mathrm{C}_{2}\right)},
\end{aligned}
\end{equation}
where, $x$ and $y$ represent the pixel position, $\text{MAX}_{I}=255$ and $\text{MSE}=\text{RMSE}^2$, $\mu_{x}$, $\mu_{y}$, $\sigma_{x}^2$ and $\sigma_{y}^2$ independently denote the pixel sample mean and the variance of $x$ and $y$, $\sigma_{xy}$ denote the cross-correlation of $x$ and $y$. Additionally $\mathrm{C}_1 = 2.55^2$ and $\mathrm{C}_2 = 7.65^2$. We calculate the local SSIM within patch of size 7×7. Subsequently, averaging all local SSIM yields the SSIM between the two aggregated edge-modulated event frames.

\begin{table*}[htbp]
\vspace{-0.0em}
\begin{center}
\caption{Left-right consistency ablation. Sequential removals of the left-right consistency census loss and EAA modules are performed from the EV-MGDispNet. Best in bold, runner-up underlined, higher values denote better performance.}
\label{tab:4}
\begin{tabular}{c|cccc|cccc}
\hline
\multirow{2}{*}{left-right consistency ablation}
&\multicolumn{4}{c|}{PSNR$\uparrow$}
&\multicolumn{4}{c}{SSIM$\uparrow$}\\
\cline{2-9} 
&\multicolumn{1}{c|}{Inter}
&\multicolumn{1}{c|}{Thun}
&\multicolumn{1}{c|}{City}
&All
&\multicolumn{1}{c|}{Inter}
&\multicolumn{1}{c|}{Thun}
&\multicolumn{1}{c|}{City}
&All\\
\hline
\multicolumn{1}{l|}{EV-MGDispNet (Ours)}
&\multicolumn{1}{c|}{\textbf{23.0037}}
&\multicolumn{1}{c|}{\textbf{20.9327}}
&\multicolumn{1}{c|}{\textbf{21.5310}}
&\textbf{21.9630}
&\multicolumn{1}{c|}{\textbf{0.9392}}
&\multicolumn{1}{c|}{\textbf{0.9172}}
&\multicolumn{1}{c|}{\textbf{0.9181}}
&\textbf{0.9244}\\
\hline
\multicolumn{1}{l|}{\quad w/o census loss}
&\multicolumn{1}{c|}{\underline{22.9646}}
&\multicolumn{1}{c|}{\underline{20.9239}}
&\multicolumn{1}{c|}{\underline{21.4901}}
&\underline{21.9234}
&\multicolumn{1}{c|}{\underline{0.9390}}
&\multicolumn{1}{c|}{\underline{0.9171}}
&\multicolumn{1}{c|}{\underline{0.9179}}
&\underline{0.9243} \\
\hline
\multicolumn{1}{l|}{\quad w/o EAA, census loss}
&\multicolumn{1}{c|}{22.5188}
&\multicolumn{1}{c|}{20.5165}
&\multicolumn{1}{c|}{21.2384}
&21.6090
&\multicolumn{1}{c|}{0.9363}
&\multicolumn{1}{c|}{0.9129}
&\multicolumn{1}{c|}{0.9156}
&0.9218\\
\hline
\end{tabular}
\vspace{-1.0em}
\end{center}
\end{table*}

\begin{figure}[htbp]
\vspace{-0.0em}
\centering
\includegraphics[width=1\columnwidth]{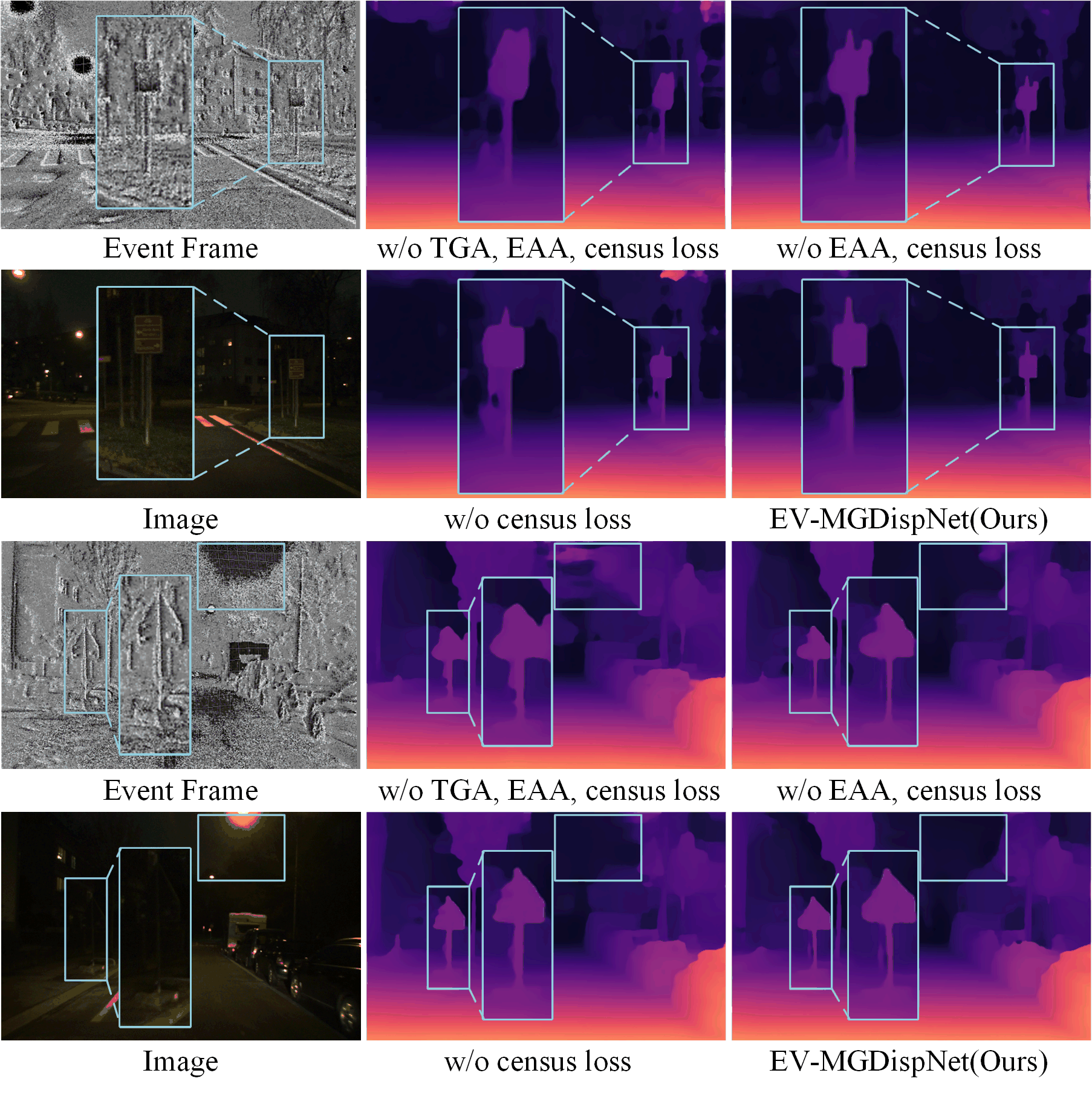}
\vspace{-1.5em}
\caption{Night scene ablation. Visual comparison of night scene between the EV-MGDispNet and the models with removed MGA Module, EAA Module and left-right consistency census loss.}
\label{fig:6}
\vspace{-1.0em}
\end{figure}

Through the analysis of experimental data, we examine the effects of EAA module, MGA module and left-right consistency census loss. A detailed analysis is as follows:

\subsubsection{Impact of EAA module}
\label{Impact of SpadeUnet}
As demonstrated in Table \ref{tab:2} and Table \ref{tab:3}, the comparison between ``w/o EAA, census loss" and ``w/o census loss" illustrates that ablating the EAA module results in a degradation of multiple metrics on both the validation set and night scenes.
In Table \ref{tab:4}, a similar comparison further demonstrates that ablating the EAA module leads to a decrease in the left-right consistency of event representations. Moreover, the integration of observations from Table \ref{tab:2} and Table \ref{tab:3} reveals that improved left-right consistency of event representations often corresponds to better performance in stereo disparity estimation metrics.
In the comparison presented in Fig. \ref{fig:5}, it is evident that after ablating the EAA module, Fig. \ref{fig:5_3} exhibits larger errors in scene structures such as power lines, vehicle outlines, grassy areas and barriers compared to Fig. \ref{fig:5_4}. This illustrates that the aggregated edge-modulated event frames generated by the EAA module are advantageous for stereo disparity estimation of fine structures.
In the night scene depicted in Fig. \ref{fig:6}, it is evident that compared to ``w/o census loss" , ``w/o EAA, census loss" reveals more inaccurate outlines of the traffic signs due to the ablation of the EAA module.
These experimental results indicate that the aggregated edge-modulated event frames generated by the EAA module provide clearer event representations, as they exhibit higher left-right consistency. Moreover, this clearer representation improves stereo disparity estimation performance.

\subsubsection{Impact of MGA}
In Table \ref{tab:2} and Table \ref{tab:3}, the comparison between ``w/o MGA, EAA, census loss" and ``w/o EAA, census loss" demonstrates that ablating the MGA module results in a degradation of multiple metrics on both the validation set and night scenes.
In the contrast depicted in Fig. \ref{fig:5}, it is clearly observable that the ablation of the MGA module results in considerably higher stereo disparity estimation errors in grassy regions and fine structures, as depicted in Fig. \ref{fig:5_2} compared to Fig. \ref{fig:5_3}. This demonstrates that the MGA module effectively enhances stereo disparity estimation results at scene edges, object contours and plane locations.
In the night scene depicted in Fig. \ref{fig:6}, it is evident that ``w/o MGA, EAA, census loss" exhibits noticeably poorer outlines of the traffic signs compared to ``w/o EAA, census loss" Additionally, the estimation at the locations of streetlights is significantly affected by the intense flickering of nighttime flickering light regions, leading to confused disparity outcomes, whereas the MGA module can alleviate this influence.
These experimental results indicate that the MGA module, guided by motion confidence maps, generates edge-enhanced feature maps with fewer pixel shifts, which in turn facilitate the construction of higher-quality cost volumes and thereby improve stereo disparity estimation results.
\subsubsection{Impact of left-right consistency census loss}
The comparison between ``w/o census loss" and ``EV-MGDispNet (Ours)" in Table \ref{tab:2} and Table \ref{tab:3} reveals that the ablation of the left-right consistency census loss leads to a deterioration in multiple metrics on both the validation set and night scenes.
In Table \ref{tab:4}, the same comparison underscores that the ablation of left-right consistency census loss results in diminished left-right consistency of event representations.
In the comparison illustrated in Fig. \ref{fig:5}, it is apparent that the ablation of left-right consistency census loss, as depicted in Fig. \ref{fig:5_4} relative to Fig. \ref{fig:5_5}, leads to amplified stereo disparity estimation errors in scene planes and fine structures. This suggests that the direct supervision of the aggregated edge-modulated event frame by the left-right consistency census loss indirectly improves stereo disparity estimation results.
In Fig. \ref{fig:6}, within the night scene, a clear distinction is observable between ``w/o census loss" and ``EV-MGDispNet (Ours)" indicating that the ablation of left-right consistency census loss leads to inferior outlines of the traffic signs.
These experimental results demonstrate that left-right consistency census loss can enhance the left-right consistency of event representations. Moreover, this improvement in left-right consistency results in event representations that more closely adhere to the stereo camera model, thereby enhancing the precision of stereo disparity estimation.

\section{Conclusions}
\label{sec:conclusions}

In this article, we scrutinize the limitations of existing methods from the perspective of network architecture and loss function.
Building upon this analysis, we propose a novel motion-guided event-based stereo disparity estimation network, named EV-MGDispNet, which integrates three key steps. 
First, we propose the edge-aware aggregation (EAA) module, which utilizes a SPADE upsampling module to modulate the distribution of feature maps using motion confidence maps, thereby generating a new clear event representation termed aggregated edge-modulated event frame.
Second, we propose a motion-guided attention (MGA) module, which leverages motion confidence maps to guide event representation features before constructing cost volumes, enhancing edge features to produce edge-enhanced feature maps with fewer pixel shifts. This facilitates the construction of high-quality cost volumes, thereby improving stereo disparity estimation performance.
In summary, we fully integrate previously overlooked temporal information into the network architecture through motion confidence maps.
Additionally, we not only use the smooth $L_{1}$ loss function but also add a census left-right consistency loss function to enhance the supervision of the EAA module. This enhancement not only strengthens the left-right consistency of event representations but also aligns them more closely with the stereo camera model, thereby enhancing the accuracy of stereo disparity estimation.
Experimental results on the DSEC dataset demonstrate that our method outperforms all known methods to date.
In the future, we will explore frame-event fusion methods to achieve higher performance\cite{hou2023fe}, as well as deep spike-based stereo disparity estimation network architectures suitable for energy-efficient inference\cite{jiang2023neuro,hu2024spike}.

\bibliographystyle{IEEEtran}
\bibliography{mybibfile}

\begin{IEEEbiography}[{\includegraphics[width=1in,height=1.25in,clip,keepaspectratio]{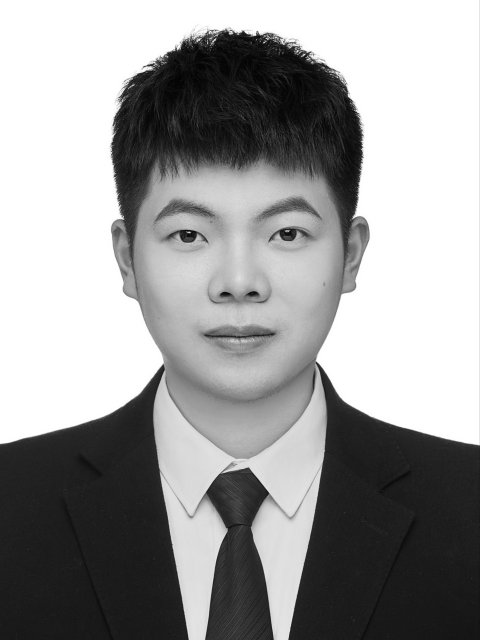}}]{Junjie Jiang}
received the B.S. degree in automation from Northeastern University at Qinhuangdao, Qinhuangdao, China, in 2020 and the M.S. degree in robot science and engineering from Northeastern University, Shenyang, China, in 2023. He is currently pursuing the Ph.D. degree in robot science and engineering with Northeastern University, Shenyang, China. His research interests include event-based vision, spiking neural network, robot visual navigation, and reinforcement learning.\end{IEEEbiography}
\vspace{-3em}
\begin{IEEEbiography}[{\includegraphics[width=1in,height=1.25in,clip,keepaspectratio]{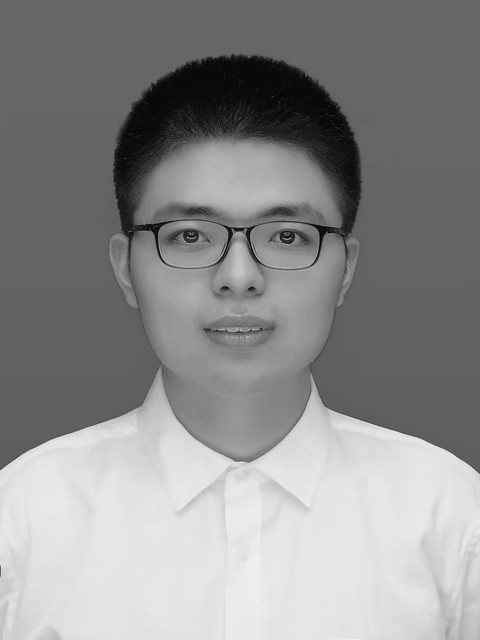}}]{Hao Zhuang}
received the B.S. degree in mechanical and electronic engineering from the Chongqing University of Posts and Telecommunications, Chongqing, China, in 2021. He is currently pursuing the M.S. degree in control engineering with Northeastern University, Shenyang, China. His research interests include event-based vision, optical flow estimation, and deep learning.\end{IEEEbiography}
\vspace{-3em}
\begin{IEEEbiography}[{\includegraphics[width=1in,height=1.25in,clip,keepaspectratio]{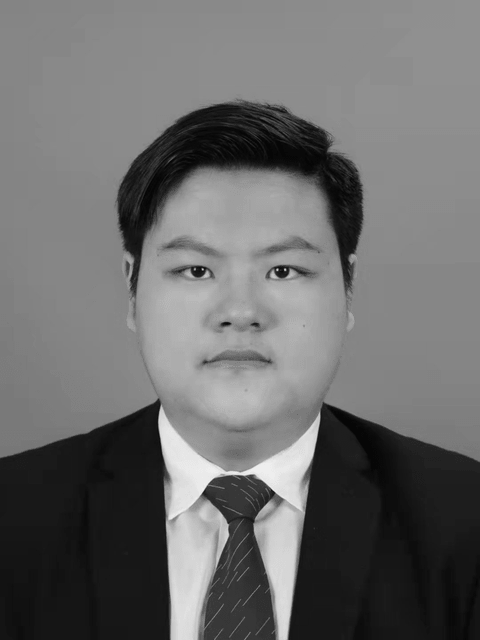}}]{Xinjie Huang}
received the B.S. degree in automation from the Zhejiang University of Technology, Hangzhou, China, in 2021. He is currently pursuing the M.S. degree in robot science and engineering with Northeastern University, Shenyang, China. His research interests include event-based vision, depth estimation, and deep learning.\end{IEEEbiography}
\vspace{-3em}
\begin{IEEEbiography}[{\includegraphics[width=1in,height=1.25in,clip,keepaspectratio]{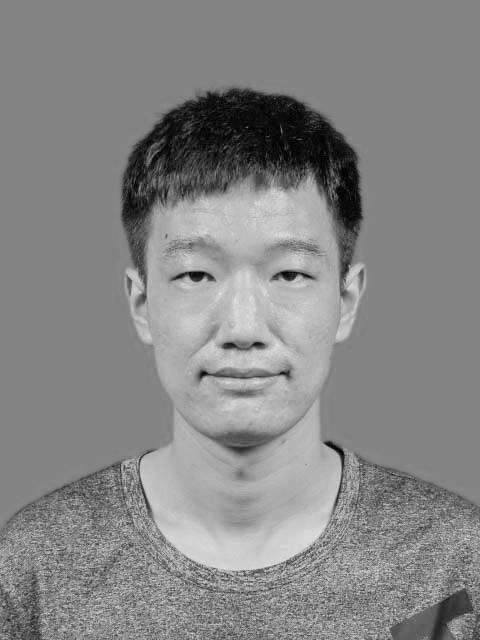}}]{Delei Kong}
(Graduate Student Member, IEEE) received the B.S. degree in automation from Henan Polytechnic University, Zhengzhou, China, in 2018, and the M.S. degree in control engineering from Northeastern University, Shenyang, China, in 2021. From 2020 to 2021, he was an Algorithm Intern with SynSense Tech. Co. Ltd., Chengdu, China. From 2021 to 2022, he was an R\&D Engineer (Advanced Vision) with the Machine Intelligence Laboratory, China Nanhu Academy of Electronics and Information Technology (CNAEIT), Jiaxing, China. Since 2022, he has been a Ph.D. student in control science and engineering from Hunan University, Changsha, China. His research interests include event-based vision, robot visual navigation, and neuromorphic computing.\end{IEEEbiography}
\vspace{-3em}
\begin{IEEEbiography}[{\includegraphics[width=1in,height=1.25in,clip,keepaspectratio]{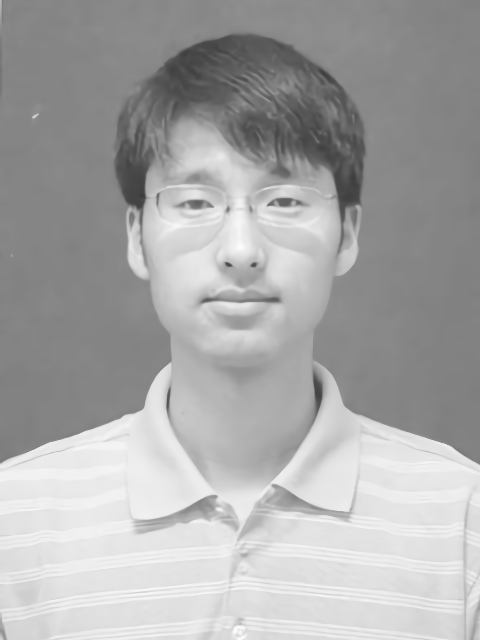}}]{Zheng Fang}
(Member, IEEE) received the B.S. degree in automation and the Ph.D. degree in pattern recognition and intelligent systems from Northeastern University, Shenyang, China, in 2002 and 2006, respectively. He was a Post-Doctoral Research Fellow with the Robotics Institute, Carnegie Mellon University (CMU), Pittsburgh, PA, USA, from 2013 to 2015. He is currently a Professor with the Faculty of Robot Science and Engineering, Northeastern University. He has published over 80 papers in well-known journals or conferences in robotics and computer vision. His research interests include visual/laser simultaneous localization and mapping (SLAM), and perception and autonomous navigation of various mobile robots.\end{IEEEbiography}

\end{document}